\documentclass[journal,twoside,web]{ieeecolor}
\usepackage{tmi}
\usepackage{cite}
\usepackage{amsmath,amssymb,amsfonts}
\usepackage{algorithmic}
\usepackage{graphicx}
\usepackage{textcomp}

\usepackage{url}% cz add
\usepackage{color}% cz add
\usepackage{subfigure}% cz add
\usepackage{graphicx}% cz add
\usepackage{multirow}% cz add
\usepackage{booktabs} % cz add
\usepackage[marginal]{footmisc} % cz add
\usepackage{tipa} % cz add
\usepackage{float} % cz add
\usepackage{placeins}
\usepackage{bm} % cz add
\usepackage{array} % cz add
\usepackage{enumerate}% cz add 20190813
\usepackage{hyperref}
\usepackage{pifont} % xq add
\usepackage{subfloat}
\usepackage{subfigure}
\usepackage{amsmath,bm}%  cz add 20200511

\usepackage{diagbox}
\usepackage{booktabs}
\usepackage{amssymb}% cz add
\usepackage{balance} % cz add
\usepackage{bbm} % cz add

\usepackage{amsmath}% cz add

\def\BibTeX{{\rm B\kern-.05em{\sc i\kern-.025em b}\kern-.08em
    T\kern-.1667em\lower.7ex\hbox{E}\kern-.125emX}}
\begin{document}
% \title{Prompt-Free Segmentation Framework for Universal Histopathology Nuclei Images}
\title{Co-Seg++: Mutual Prompt-Guided Collaborative Learning for Versatile Medical Segmentation}
\author{Qing Xu, Yuxiang Luo, Wenting Duan, Zhen Chen
        \thanks{\quad This work was supported in part by the Program of China Scholarship Council (202508330191). \textit{(Corresponding author: Zhen Chen)}} 
        \thanks{\quad Q. Xu is with School of Computer Science, University of Nottingham Ningbo China, Ningbo, Zhejiang, China, and with University of Nottingham, UK (e-mail: qing.xu@nottingham.edu.cn).}
        \thanks{\quad Y. Luo is with the Graduate School of Information, Production and Systems, Waseda University, Japan (e-mail: yuxiang.luo@ruri.waseda.jp).}
        \thanks{\quad W. Duan is with School of Engineering and Physical Science, University of Lincoln, Lincoln  LN6 7TS, UK (email: wduan@lincoln.ac.uk).}
        \thanks{\quad Z. Chen is with Yale University, New Haven, CT 06510, USA (e-mail: zchen.francis@gmail.com).}
        % \thanks{\quad \textit{(Corresponding author: Zhen Chen)}}
        }

\maketitle
\begin{abstract}
Medical image analysis is critical yet challenged by the need of jointly segmenting organs or tissues, and numerous instances for anatomical structures and tumor microenvironment analysis. Existing studies typically formulated different segmentation tasks in isolation, which overlooks the fundamental interdependencies between these tasks, leading to suboptimal segmentation performance and insufficient medical image understanding. To address this issue, we propose a Co-Seg++ framework for versatile medical segmentation. Specifically, we introduce a novel co-segmentation paradigm, allowing semantic and instance segmentation tasks to mutually enhance each other. We first devise a spatio-sequential prompt encoder (SSP-Encoder) to capture long-range spatial and sequential relationships between segmentation regions and image embeddings as prior spatial constraints. Moreover, we devise a multi-task collaborative decoder (MTC-Decoder) that leverages cross-guidance to strengthen the contextual consistency of both tasks, jointly computing semantic and instance segmentation masks. Extensive experiments on diverse CT and histopathology datasets demonstrate that the proposed Co-Seg++ outperforms state-of-the-arts in the semantic, instance, and panoptic segmentation of dental anatomical structures, histopathology tissues, and nuclei instances. The source code is available at \url{https://github.com/xq141839/Co-Seg-Plus}.
\end{abstract}

\begin{IEEEkeywords}
Collaborative Learning, medical semantic segmentation, medical instance segmentation, mutual prompt
\end{IEEEkeywords}

\section{Introduction}
\label{sec:introduction}

\IEEEPARstart{M}{edical} image segmentation plays a crucial role in clinical applications and has received significant research attention \cite{shui2024unleashing,horst2024cellvit,cheng2024unleashing}. For example, histopathology image analysis demands accurate segmentation of tissue regions and separation of individual nuclei within these regions to assess tissue subtypes and tumor grading \cite{zhao2024foundation}. Similarly, cone beam computed tomography (CBCT) requires precise delineation of anatomical structures such as jaw bones and sinuses, alongside instance-level segmentation of individual teeth for dental treatment planning \cite{bolelli2024segmenting}. These dual requirements have given rise to the challenging field of versatile medical segmentation, which seeks to unify both semantic and instance segmentation tasks across various medical imaging contexts.

\begin{figure}[!t]
  % \vspace{-4.3cm}
  \centering
  \includegraphics[width=1\linewidth]{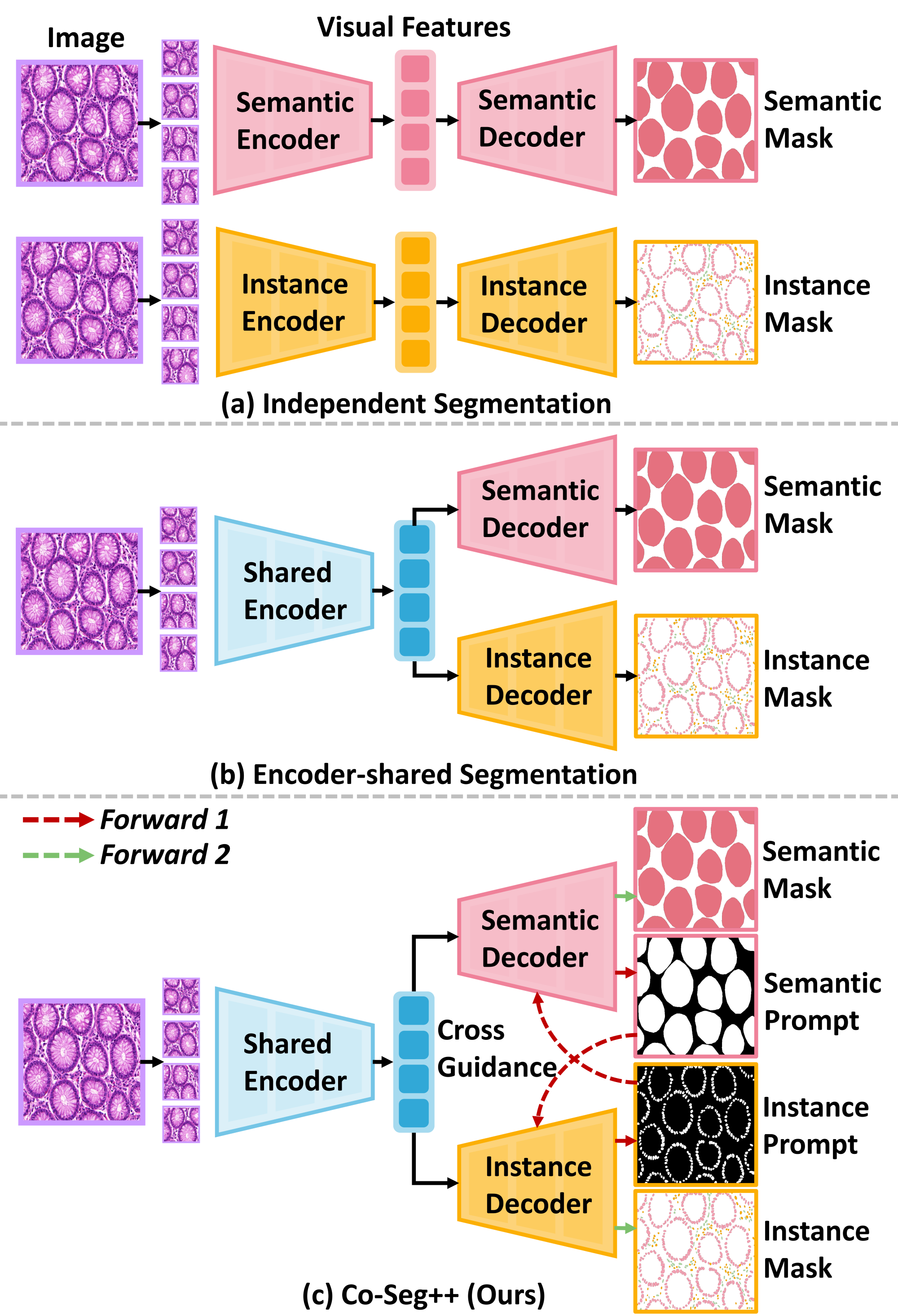}
  \caption{Comparison of our Co-Seg++ and existing segmentation works. (a) Two independent networks for semantic and instance segmentation. (b) A shared image encoder but separated task decoders for semantic and instance segmentation. (c) Our Co-Seg++ leverages spatio-temporal prompts for collaborative semantic and instance segmentation.}
  \label{fig:intro}
  % \vspace{-1.0em}
\end{figure}

\begin{figure}[!t]
  % \vspace{-4.3cm}
  \centering
  \includegraphics[width=0.7\linewidth]{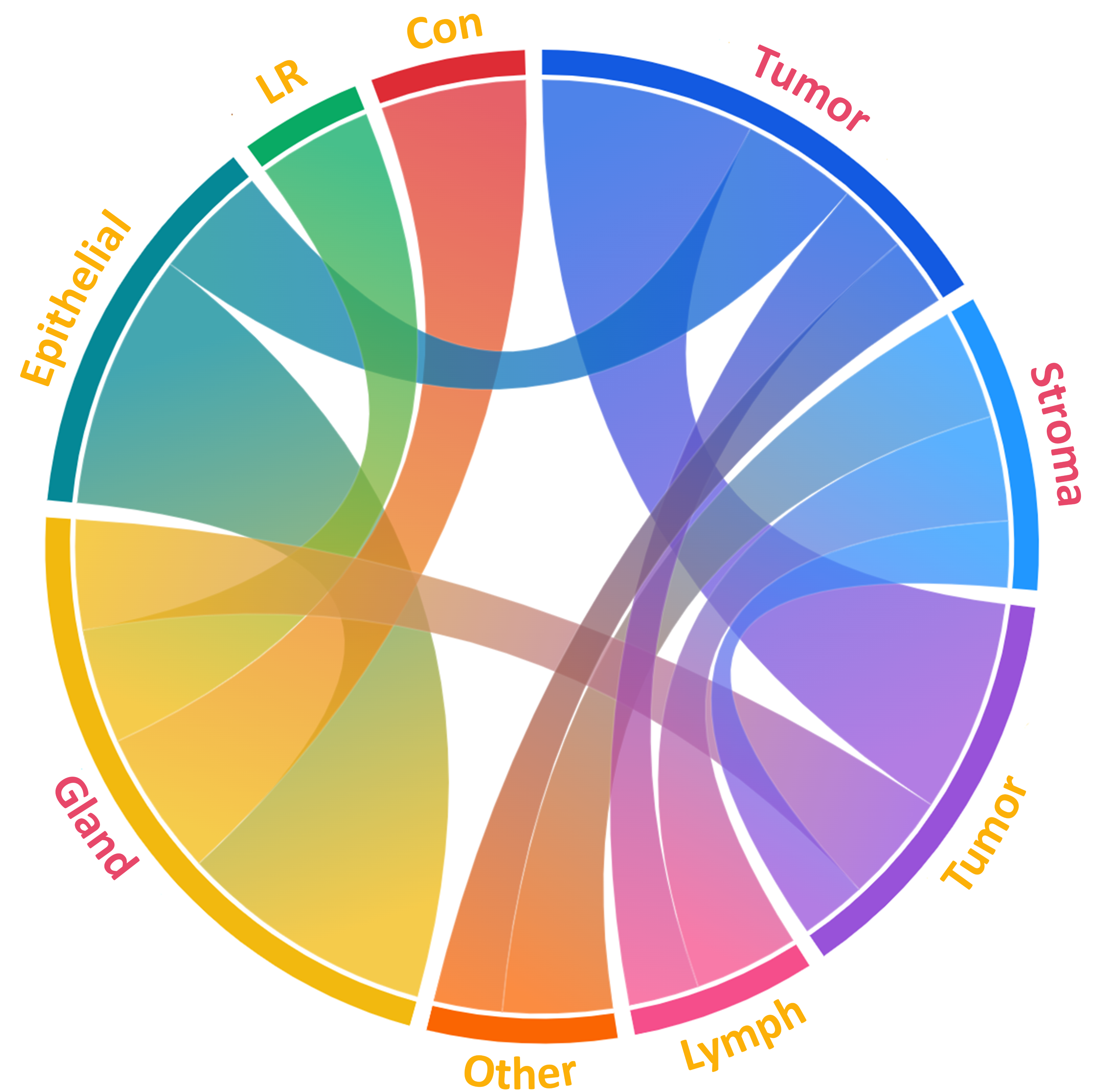}
  \caption{Regional interdependencies between 
  semantic and instance segmentation tasks in nuclei images. Semantic regions (\textit{e.g.}, glands) often encompass multiple types of instance-level structures (\textit{e.g.}, lympho-reticular (LR) and connective (Con) nuclei). The relationship of semantic and instance segmentation confirms contextual understanding and motivates the proposed joint optimization of versatile medical segmentation.}
  \label{fig:interdependencies}
  % \vspace{-1.0em}
\end{figure}

Typically, medical image segmentation frameworks \cite{ronneberger2015u, isensee2021nnu, rahman2024emcad} have been designed for specific tasks, resulting in isolated solutions that address semantic segmentation and instance segmentation separately, as shown in Fig. \ref{fig:intro}(a). These conventional approaches typically adopt independent encoder-decoder architectures without shared information during the training process. The semantic segmentation networks are optimized to classify each pixel into pre-defined categories, meanwhile instance segmentation models \cite{graham2019hover, chen2023cpp, horst2024cellvit} focus on detecting and delineating individual objects. However, multiple independent networks \cite{stringer2021cellpose, liu2024swin, lou2025nusegdg} lead to increased computation complexity. More seriously, this separation learning \cite{pocock2022tiatoolbox} impairs consistency for similar low-level features across tasks, resulting in redundant feature extraction. These limitations motivate the development of more integrated approaches that could leverage commonalities between semantic and instance segmentation tasks in medical images.

To this end, the encoder-shared framework has emerged as a promising alternative paradigm for versatile medical segmentation, as shown in Fig. \ref{fig:intro}(b). These methods \cite{graham2023one, griebel2025segment} utilize a common feature extraction backbone for both semantic and instance segmentation tasks, enabling more efficient resource utilization and establishing a foundation for potential information sharing. The shared encoder captures universal representations beneficial to both tasks, reducing computational redundancy and providing consistency in the initial feature extraction stage. While this paradigm improves efficiency compared to fully independent models, it still maintains separate decoders, leading to limited cross-task interaction during the critical decoding phase, which fails to exploit the mutual guidance that could be derived from intermediate representations of each task. In fact, we observe that both segmentation tasks are highly correlated, by analyzing the regional interdependencies of multiple histopathology datasets \cite{giaf011, sirinukunwattana2017gland, graham2023one} in Fig. \ref{fig:interdependencies}. These segmentation tasks aim to achieve an adequate perception and understanding of histopathology images. For example, accurately identifying nuclei provides valuable cues for understanding the underlying tissue structures, whereas tissue segmentation can aid in localizing nuclei. These synergies motivate us to develop a collaborative framework that jointly optimizes semantic and instance segmentation to improve medical image analysis.

To overcome this bottleneck, we propose Co-Seg++, a collaborative versatile medical segmentation framework that allows semantic and instance segmentation tasks to mutually enhance each other. As illustrated in Fig. \ref{fig:intro}(c), Co-Seg++, based on a novel co-segmentation paradigm, improves segmentation mask quality by capturing contextual dependencies between the two tasks. Specifically, we design a spatio-sequential prompt encoder (SSP-Encoder) that adaptively integrates spatial context and sequential dynamics to generate prior spatial constraints for guiding segmentation decoding across both tasks. We further introduce a multi-task collaborative decoder (MTC-Decoder) that leverages cross-task guidance to jointly predict semantic and instance segmentation masks in a unified framework. This collaborative learning strategy not only achieves enhanced contextual consistency but also significantly reduces prediction errors through mutual task optimization. Extensive experiments on diverse histopathology and dental datasets demonstrate that our Co-Seg++ consistently outperforms state-of-the-art segmentation methods.

The contributions of this work are summarized as follows:
\begin{itemize}
    \item We propose Co-Seg++, a versatile medical segmentation framework that incorporates a novel co-segmentation paradigm, enabling semantic and instance segmentation tasks to mutually enhance each other. 
    \item We devise a SSP-Encoder that integrates long-range spatial and sequential information to extract high-quality semantic and instance prompts, providing effective prior knowledge for cross-task segmentation guidance.
    \item We design a MTC-Decoder that employs shared image embeddings and joint prompt mechanisms for collaborative semantic and instance segmentation decoding, achieving mutual task optimization and enhanced mask prediction consistency. 
    \item We conduct extensive experiments on diverse medical image
    datasets, and our Co-Seg++ framework outperforms state-of-the-art independent and encoder-shared medical segmentation methods across semantic, instance, and panoptic segmentation tasks. 
\end{itemize}

\begin{figure*}[!t]
  % \vspace{-4.3cm}
  \centering
  \includegraphics[width=0.9\linewidth]{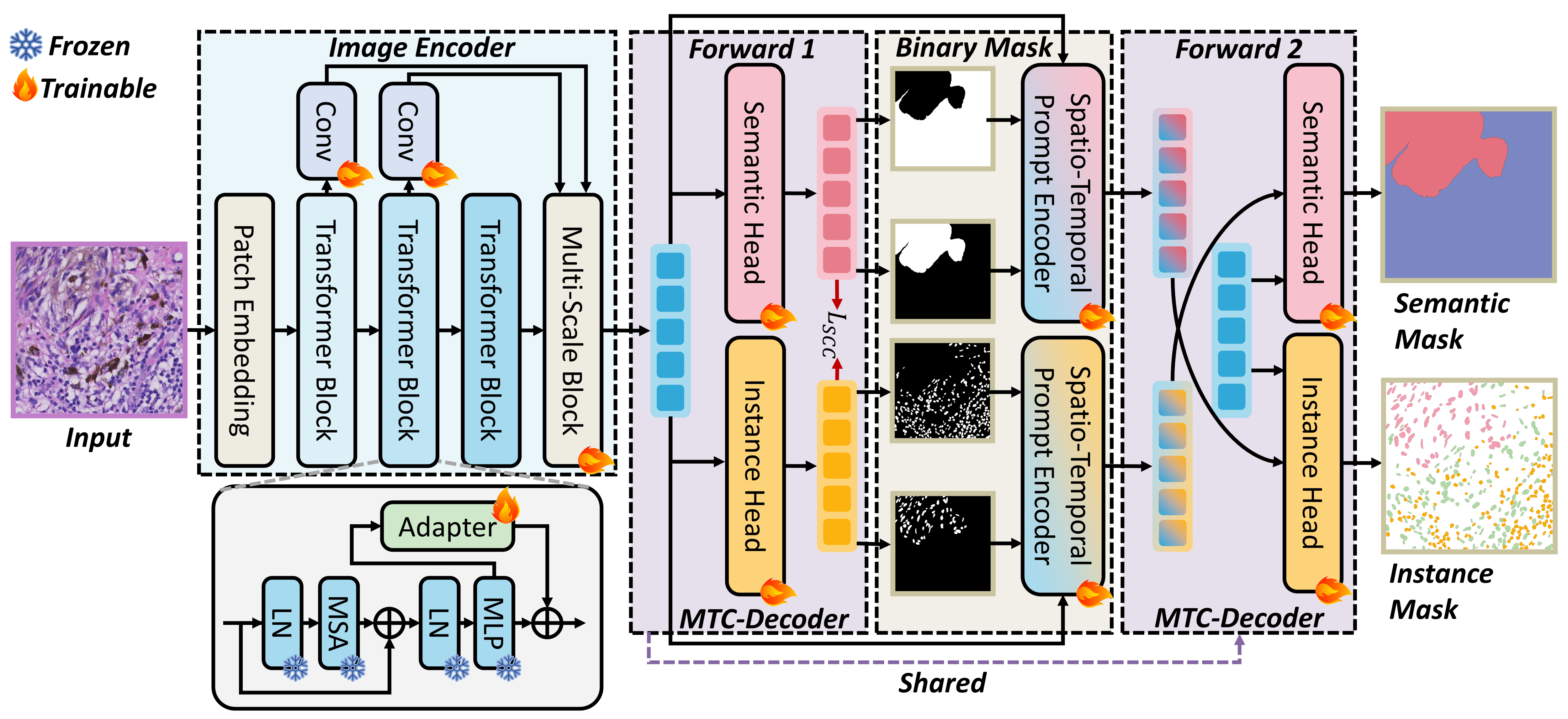}
  \caption{The overview of our Co-Seg++ framework for collaborative versatile medical segmentation, consisting of SSP-Encoder that captures long-range spatial and sequential relationships and MTC-Decoder with semantic and instance heads for mutual task guidance. Co-Seg++ operates through two forward passes: the first pass generates binary masks from both heads as prior spatial constraints supervised by a spatial consistency constraint loss. In the second forward pass, the semantic and instance heads in MTC-Decoder leverage these constraints via cross-attention to mutually enhance semantic region delineation and instance classification. The Co-Seg++ framework fully exploits complementary contextual information through spatio-temporal prompts.}
  \label{fig:method}
  % \vspace{-1.0em}
\end{figure*}

\section{Related Work}
\subsection{Medical Image Segmentation}
Medical image segmentation is fundamental to computer-aided diagnosis and treatment planning \cite{ronneberger2015u, isensee2021nnu, rahman2024emcad}. This field encompasses semantic segmentation (\textit{i.e.}, pixel-level classification), instance segmentation (\textit{i.e.}, individual object delineation), and panoptic segmentation that unifies both by jointly predicting semantic classes and instance identities \cite{kirillov2019panoptic}. Building on the U-Net encoder-decoder framework \cite{ronneberger2015u}, recent advances span three network architectures: convolutional neural network (CNN), vision transformers (ViT) \cite{dosovitskiy2020image}, and more recent Mamba models \cite{liu2024vmamba}. The CNN-based methods \cite{isensee2021nnu, xu2023dcsau, rahman2024emcad} applied automated architecture optimization and multi-scale convolutions to enhance target awareness. The ViT-based methods \cite{chen2024transunet, hao2024t} leveraged self-attention mechanisms to capture global contexts. The Mamba-based methods \cite{liu2024swin, cheng2025mamba} utilized sequence modeling to handle long-range relationships with linear computational complexity. These works have demonstrated strong adaptability across diverse imaging modalities.

Histopathology and CBCT represent key modalities for tumor assessment and dental practices, with applications spanning semantic and instance segmentation. In histopathology, semantic segmentation localizes tissue regions while instance segmentation identifies individual nuclei for tumor microenvironment profiling \cite{graham2019hover, raza2019micro, stringer2021cellpose}. Methods like CPP-Net \cite{chen2023cpp} used multi-point sampling for polygon distance mapping, and CellViT \cite{horst2024cellvit} employed multi-task learning for multi-directional distance maps. For CBCT analysis, semantic segmentation parses orofacial structures while instance segmentation recognizes individual teeth \cite{cui2019toothnet}. PadUNet \cite{cipriano2022improving} achieved 3D nerve canal segmentation via sparse annotations, while TSG-GCN \cite{liu2024individual} optimized spatial tooth distribution through dynamic adjacency learning. However, existing approaches employ task-specific models. Our Co-Seg++ transcends these limitations, achieving versatile segmentation with superior performance.

% Furthermore, existing medical image segmentation tasks can be categorized into semantic segmentation and instance segmentation. In particular, histopathology and CBCT imaging scenarios inherently require these segmentation capabilities. 

% histopathology and CBCT are two important modalities for tumor
% grade assessment and anatomical structure analysis, respectively.

\subsection{Versatile Segmentation Paradigms}
The classical versatile medical segmentation frameworks have predominantly adopted a task-decoupled paradigm to address multiple segmentation objectives simultaneously. The TIAToolbox \cite{pocock2022tiatoolbox} employed an ensemble approach that integrates multiple independent networks for tissue and nuclei segmentation in histopathology images, treating each task as a separate optimization problem. To enhance cross-task consistency, Cerberus \cite{graham2023one} employed the shared encoder with decoupled decoders for joint training efficiency. Federico et al. \cite{bolelli2025segmenting} adopted multi-class strategies with independent labels for CBCT tasks. VerSemi \cite{zeng2025segment} introduced the task-prompted dynamic learning to provide the improved decoding workflow for different segmentation tasks. However, they lack flexible guidance mechanisms that limit their ability to incorporate contextual knowledge or interactive signals, highlighting the need for more adaptive and guidance-driven segmentation paradigms.

The segment anything model (SAM) \cite{kirillov2023_sam} introduced prompt-based segmentation through point, box, text, or mask guidance. Recent studies \cite{wang2024seganypath, chen2025segment} have demonstrated the strong adaptability of SAM across a variety of medical scenarios. These prompt-driven methods enabled flexible segmentation by incorporating domain knowledge directly into the inference process. To further refine segmentation outputs, PromptNucSeg \cite{shui2024unleashing} utilized an anchor-based detection prompter to automatically produce multi-point prompts within each nucleus for deeply guiding nuclei segmentation. UN-SAM \cite{chen2025sam} devised a multi-scale self-prompting strategy to extract coarse mask prompts from image embeddings, eliminating the demand for labor-intensive manual annotations. However, these prompt-based approaches are designed to improve performance within individual segmentation tasks. Unlike existing approaches that utilize prompts for task-specific enhancement, Co-Seg++ explores complementarities between semantic and instance segmentation in medical scenarios, which enables both tasks to mutually guide each other.

\section{Methodology}

We present the Co-Seg++ framework with the co-segmentation paradigm in Fig. \ref{fig:method}, and achieve the mutual optimization between semantic and instance segmentation for medical images. To accomplish this, we design the SSP-Encoder to provide prior spatial constraints to guide both tasks, and the MTC-Decoder to cooperatively generate semantic and instance maps through bidirectional information interactions, which enables semantic and instance segmentation tasks to enhance each other for improved segmentation accuracy.

\begin{figure}[!t]
  % \vspace{-4.3cm}
  \centering
  \includegraphics[width=0.9\linewidth]{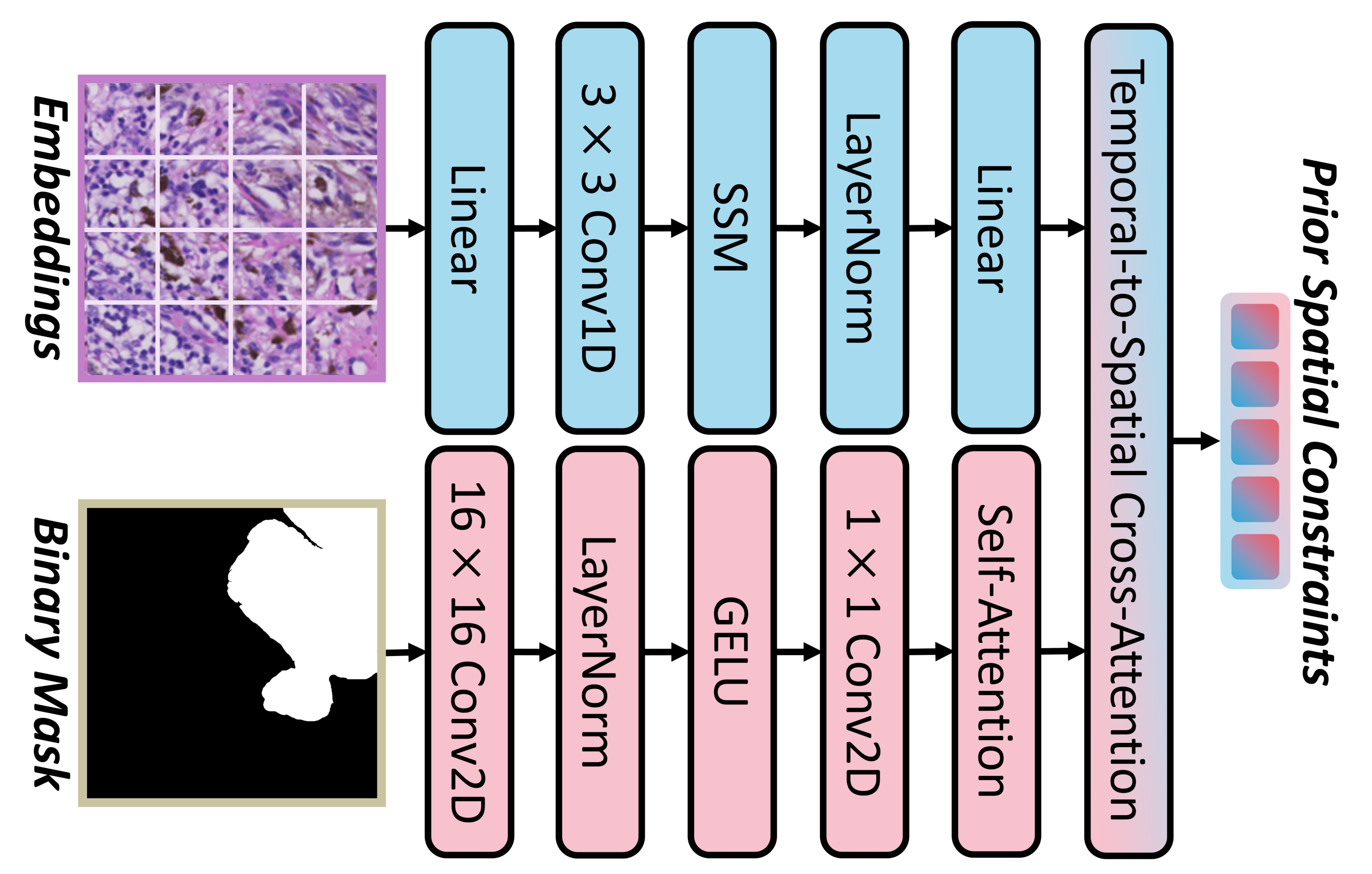}
  \caption{The illustration of SSP-Encoder, integrating spatial prompts and sequential memory to establish long-range relationships between target segmentation regions and shared image embeddings, providing prior spatial constraints for cross-task guidance.}
  \label{fig:prompt}
  % \vspace{-1.0em}
\end{figure}

\subsection{Co-Segmentation Paradigm}
Existing versatile medical segmentation methods \cite{graham2023one, griebel2025segment} decouple the parameter space of semantic and instance segmentation, disrupting their interdependencies, degrading the performance. To address this issue, we propose the co-segmentation paradigm that leverages the closed-loop bidirectional interaction to realize dual-task collaborative optimization. Specifically, our goal is to train a model $f_\Theta: x \rightarrow \{y_{1}, y_{2}\}$, where $\Theta=\{\theta_{1}, \theta_{2}\}$ represents the parameters for semantic segmentation $\theta_1$ and instance segmentation $\theta_2$, with $y_{1}$ and $y_{2}$ as the corresponding segmentation masks. To model the interdependence between the two tasks, we define their joint prediction of $y_1$ and $y_2$ as follows:
\begin{equation} \label{eq:1}
\begin{split}
p_{\theta_1, \theta_2}(y_1, y_2 | x) &= p_{\theta_1}(y_1 | x) \cdot p_{\theta_2}(y_2 | y_1, x) \\ &= p_{\theta_2}(y_2 | x) \cdot p_{\theta_1}(y_1 | y_2, x).
\end{split}
\end{equation}
Note that Eq. \eqref{eq:1} reveals the mutual dependency between semantic and instance segmentation through intertwined conditional probabilities. It states that $p_{\theta_1, \theta_2}(y_1, y_2 | x)$, the joint probability of obtaining both segmentations given the image and model parameters, is decomposed in two symmetrical ways:
\begin{itemize}
  \item $p_{\theta_1}(y_1 | x) \cdot p_{\theta_2}(y_2 | y_1, x)$ suggests that once the semantic segmentation $y_1$ is known, it directly influences the prediction of the instance segmentation $y_2$.
  \item Similarly, $p_{\theta_2}(y_2 | x) \cdot p_{\theta_1}(y_1 | y_2, x)$ implies that knowing the instance segmentation $y_2$ affects the outcome of the semantic segmentation $y_1$.
\end{itemize}
Both expressions are mathematically equivalent because they describe the same joint probability from different perspectives, highlighting that each segmentation task provides crucial context that enhances the accuracy of the other. This reciprocal relationship underscores the need for a collaborative approach, where both segmentation tasks are optimized together rather than in isolation, leveraging the full context available from each task to improve overall segmentation results. The gradient calculations of the relationship for $\theta_1$ and $\theta_2$ is defined as:
\begin{equation} \label{eq:2}
\begin{split}
\nabla_{\theta_i} \ell_i = &-\nabla_{\theta_i} \mathbb{E} \left[ \log p(y_i | x, \theta_i) \right] \\ &- \nabla_{\theta_i} \mathbb{E} \left[ \log p(y_j | y_i, x, \theta_j) \right] \\ &+ \nabla_{\theta_i} I(y_1; y_2 | x),
\end{split}
\end{equation}
where $i, j \in \{1,2\}, i\neq j$ and $\nabla_{\theta} I(y_1; y_2 | x) = \mathbb{E} \left[ \nabla_{\theta} \log \frac{p(y_1, y_2 | x)}{p(y_1 | x)p(y_2 | x)} \right]$. This term captures the mutual information between 
$y_1$ and $y_2$, ensuring that both segmentation tasks benefit from shared feature learning. The first term of Eq. \eqref{eq:2} represents the main segmentation loss gradient, while the second term incorporates the implicit gradient of task interdependencies, modeling the complementarity between $\theta_1$ and $\theta_2$. Note that although $p(y_j | y_i, x, \theta_j)$ is parameterized by $\theta_j$, we include its gradient with respect to $\theta_i$ to reflect the implicit dependency introduced by co-training and shared feature representations during optimization. As such, this reflects cross-task interactions arising from shared components (\textit{e.g.}, shared encoder or joint decoder), enabling gradient flow between tasks. The third term captures the mutual information between the two outputs, ensuring both tasks benefit from shared learning signals. This leads to the following optimization update rule:
\begin{equation}
    \theta_1 \leftarrow \theta_1 - \eta \nabla_{\theta_1} \ell_1, \quad \theta_2 \leftarrow \theta_2 - \eta \nabla_{\theta_2} \ell_2,
\end{equation}
where $\eta$ is the learning rate. In this way, the co-segmentation paradigm breaks the barrier of isolating gradient flows from each other in multi-task learning, enabling our Co-Seg++ framework to achieve the mutual optimization of semantic and instance segmentation.

\begin{figure}[!t]
  % \vspace{-4.3cm}
  \centering
  \includegraphics[width=1\linewidth]{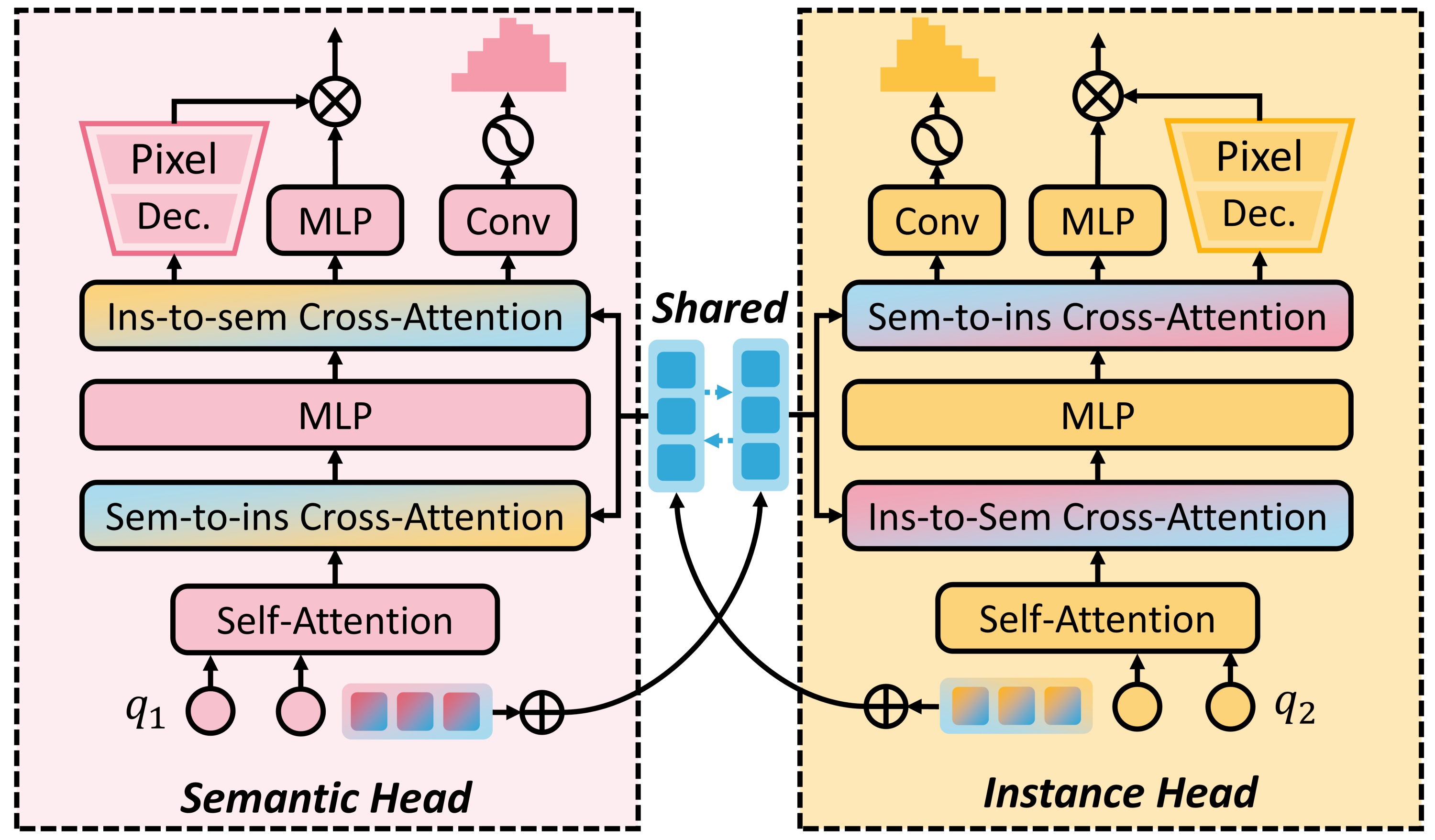}
  \caption{The architecture of the MTC-Decoder, employing cross-guidance mechanisms and probability distribution alignment to enable semantic and instance segmentation tasks to mutually enhance each other while ensuring spatial consistency in segmentation decoding.}
  \label{fig:decoder}
  % \vspace{-1.0em}
\end{figure}

\subsection{Spatio-Sequential Prompt Encoder}
The existing prompt-based segmentation methods \cite{kirillov2023_sam, ravi2025sam} typically employ two distinct types of guidance signals: the sparse prompts (\textit{e.g.}, point and box) and the dense prompts (\textit{e.g.}, masks). While the sparse prompts provide precise region-of-interest cues, they offer limited contextual information for complex segmentation tasks compared to dense prompts. The dense prompts, in contrast, provide more comprehensive spatial cues but typically lack temporal context or task-specific structure. The proposed co-segmentation paradigm requires establishing implicit interdependencies between semantic and instance segmentation tasks. To achieve this, we introduce the SSP-Encoder that leverages region masks and enriched image features to generate prior spatial constraints, effectively incorporating spatial and sequential prompts to improve the ability of the Co-Seg++ to coordinate between semantic and instance segmentation tasks.

As elaborated in Fig. \ref{fig:prompt}, SSP-Encoder comprises two parallel branches designed to capture temporal $t$ and task-specific spatial $s_i$ prompts. The temporal branch adopts shared image embeddings $h$ as input and first applies a linear layer $\rm Linear_{down}$ for down-projection. Then, a $3 \times 3$ 1D convolution followed by a state space model (SSM) extracts sequential features and models long-range relationships. Finally, we apply another linear layer $\rm Linear_{up}$ for up-projection. These operations can be summarized as follows:
\begin{equation}
t = {\rm Linear_{up}(LN(SSM(Conv1D(Linear_{down}}(h))))),
\end{equation}
where $\rm LN(\cdot)$ denotes LayerNorm. Different from the existing Mamba networks \cite{liu2024swin, hao2024t, liu2024vmamba}, this sequential branch omits SiLU activation and gating operations since the image embeddings have already been optimally processed by the image encoder, avoiding redundant feature filtering. Moreover, the spatial branch employs several 2D convolutions to transform task-specific mask logits $m_i$ to patch embeddings, followed by self-attention $\rm SA(\cdot)$ to capture global context:
\begin{equation}
s_i = {\rm SA(Conv2D_{2}(GELU(LN(Conv2D_{1}}(m_i))))),
\end{equation}
where $\rm Conv2D_{1}$ denotes a $16 \times 16$ convolution for patch tokenization, $\rm Conv2D_{2}$ is a $1 \times 1$ convolution for channel expansion. After that, the SSP-Encoder adopts cross-attention to integrate spatial prompts with sequential prompts, as follows:
\begin{equation}
c_i = {\rm CrossAttention}(Q(s_i), K(t), V(t)).
\end{equation}
On this basis, we apply SSP-Encoder to semantic and instance mask logits, effectively obtaining prior spatial constraints $c_i$ of Co-Seg++ for guiding versatile segmentation decoding.  

\subsection{Multi-Task Collaborative Decoder}
Following the co-segmentation paradigm, we propose the MTC-Decoder to enforce bidirectional interactions between semantic and instance segmentation, ensuring mutual refinement rather than treating them as isolated tasks, as illustrated in Fig. \ref{fig:decoder}. Specifically, the MTC-Decoder contains semantic and instance heads that leverage two sets of query embeddings: $q_i$ and $q_j$ to save the decoding information of both tasks. To leverage complementary effects between tasks, we first perform self-attention on each query, followed by cross-attention with prior spatial constraints $c_j$, as follows:
\begin{equation}
q'_i = {\rm CrossAttention}(Q({\rm SA}(q_i)), K(h\oplus c_j), V(h\oplus c_j)),
\end{equation}
where $\oplus$ represents the element-wise addition. When semantic features act as keys and values in the attention mechanism, while instance features serve as queries, this design allows the cross-attention operation to integrate global contextual information from semantic masks into instance-level features, enabling spatial constraints and shape priors to refine instance boundaries and reduce false positives. For example, closely packed or overlapping nuclei instances benefit from the broader spatial context provided by semantic features. After that, we apply a MLP  to  refine the query embeddings on $q'_i$, and further conduct the reverse cross-attention to generate task-specific image embeddings $g$, as follows: 
\begin{equation}
g_i = {\rm CrossAttention}(Q(h\oplus c_j), K(q'_i), V(q'_i)).
\end{equation}
In this way, the prior spatial constraints $c_j$ of instances can improve the boundary awareness of semantic segmentation, while semantic feedback provides region constraints for instance discrimination. To further enhance spatial consistency between complementary tasks, we transform $g_i$ to a category probability distribution $y_i^{\rm prob}$ using $1 \times 1$ convolution and softmax operations. We adopt Kullback-Leibler divergence to compute spatial consistency constraint loss $\mathcal{L}_{\rm SCC}$, as follows:
\begin{equation}
\mathcal{L}_{\rm SCC} = \mathbb{E}_{x \sim y_i^{\rm prob}(x)} \left[ \log_2 y_i^{\rm prob}(x) - \log_2 y_j^{\rm prob}(x) \right].
\end{equation}
Following the standard SAM \cite{kirillov2023_sam}, each segmentation head adopts a pixel decoder to upsample the refined task-specific image embedding. Finally, our MTC-Decoder generates semantic and instance segmentation predictions by performing the dot product between their unscaled image embeddings and corresponding task queries. The predictions of MTC-Decoder contain both the low-resolution binary mask $y^{\rm bin}$ and the final segmentation mask, where $y^{\rm bin}$ is supervised by binary cross-entropy loss as follows:
\begin{equation}
\mathcal{L}_{\rm BM} = -y^{\rm bin}_i\log(\hat{y}_i^{\rm bin})-y^{\rm bin}_j\log(\hat{y}_j^{\rm bin}).
\end{equation}
To optimize the final segmentation masks, we utilize the combination of cross-entropy loss and Dice loss for semantic segmentation and follow \cite{horst2024cellvit} for instance segmentation:
\begin{equation}
\begin{split}
&\mathcal{L}_{\rm SEG}^{\rm sem} = \mathcal{L}_{\rm Dice} + \mathcal{L}_{\rm CE}, \\
& \mathcal{L}_{\rm SEG}^{\rm ins} = \mathcal{L}_{\rm Dice} + \mathcal{L}_{\rm Focal} + \mathcal{L}_{\rm MSE} + \mathcal{L}_{\rm MSGE}.
\end{split}
\end{equation}
In this way, the proposed MTC-Decoder adopts the bidirectional interactions of semantic and instance segmentation information to mutually improve the quality of segmentation masks of our Co-Seg++ framework.

\begin{table*}[!t]
% \hspace{-0.6cm}
\centering
\small
\setlength\tabcolsep{5.9pt}
\caption{Comparison with state-of-the-arts on histopathology semantic segmentation.}
\scalebox{1}{\begin{tabular}{l|ccc|ccc|ccc|ccc}
\hline
& \multicolumn{6}{c|}{PUMA} & \multicolumn{3}{c|}{CRAG} & \multicolumn{3}{c}{GlaS} \\
\cline{2-13}
Methods & \multicolumn{3}{c|}{Tumor} &  \multicolumn{3}{c|}{Stroma} &  \multicolumn{3}{c|}{Gland} &  \multicolumn{3}{c}{Gland} \\
\cline{2-13}
& Dice$\uparrow$ & mIoU$\uparrow$ & HD$\downarrow$ & Dice$\uparrow$ & mIoU$\uparrow$ & HD$\downarrow$ & Dice$\uparrow$ & mIoU$\uparrow$ & HD$\downarrow$ & Dice$\uparrow$ & mIoU$\uparrow$ & HD$\downarrow$ \\
\hline
nnUnet \cite{isensee2021nnu}  & 91.18 & 85.39 & 244.71 & 53.45 & 46.43 & 244.71 &   88.61 & 85.42 & 184.71 & 89.71 & 82.81 & 148.20\\
Swin-Umamba \cite{liu2024swin}  & 88.30 & 81.04 & 272.98 & 53.78 & 42.14 & 336.21 & 86.92 & 82.58 & 244.62 & 88.49 & 80.99 & 146.84\\
EMCAD \cite{rahman2024emcad}  & 87.68 & 80.07 & 286.11 & 52.70 & 44.92 & 343.97 & 87.84 & 83.68 & 252.69 & 85.36 & 77.50 & 210.11\\
H-SAM \cite{cheng2024unleashing}  & 91.31 & 85.47 & 249.35 & 54.30 & 47.14 & 370.92 & 88.92 & 85.77 & 222.30 & 86.84 & 79.87 & 193.29\\ 
Zig-RiR \cite{chen2025zig} & 90.91 & 85.20 & 241.56 & 55.34 & 48.32 & 242.71 & 88.35 & 84.94 & 199.28 & 86.03 & 78.83 & 212.74\\
\hline
HoverNet \cite{graham2019hover}& 86.65 & 78.72 & 346.51 & 46.99 & 39.13 & 410.99 & 84.67 & 79.41 & 257.99 & 84.12 & 75.31 & 264.28 \\
HookNet \cite{van2021hooknet} & 86.84 & 79.03 & 321.59 & 47.25 & 39.44 & 395.72 & 84.89 & 79.68 & 245.83 & 84.35 & 75.58 & 251.94 \\
CellPose \cite{stringer2021cellpose}  & 85.92 & 77.58 & 358.42 & 45.73 & 37.89 & 425.67 & 83.74 & 78.12 & 271.35 & 83.21 & 74.02 & 278.59 \\
CPPNet \cite{chen2023cpp} & 87.34 & 79.78 & 312.87 & 48.56 & 40.92 & 378.24 & 85.43 & 80.34 & 231.76 & 85.67 & 76.89 & 234.15 \\
CellViT \cite{horst2024cellvit} & 90.34 & 84.12 & 261.78 & 53.67 & 46.89 & 318.45 & 88.95 & 85.23 & 195.67 & 89.48 & 82.74 & 162.35 \\
PromptNucSeg \cite{shui2024unleashing} & 91.57 & 86.08 & 228.92 & 55.21 & 48.73 & 285.63 & 89.67 & 86.49 & 178.24 & 89.85 & 83.07 & 159.89 \\
PathoSAM \cite{griebel2025segment}  & 91.78 & 86.34 & 215.31 & 54.86 & 48.15 & 292.17 & 89.28 & 85.94 & 183.45 & 89.93 & 83.68 & 156.72 \\
\hline
Co-Seg++ & \textbf{93.15} & \textbf{88.03} & \textbf{193.34} & \textbf{60.88} & \textbf{53.65} & \textbf{231.56} & \textbf{90.78} & \textbf{87.94} & \textbf{172.85} & \textbf{91.56} & \textbf{85.47} & \textbf{135.28} \\
\hline
\end{tabular}}
\label{tab:sem_histo}
\end{table*}

\begin{table*}[!t]
% \hspace{-0.6cm}
\centering
\small
\setlength\tabcolsep{7.5pt}
\caption{Comparison with state-of-the-arts on CBCT semantic segmentation.}
\scalebox{1}{\begin{tabular}{l|ccc|ccc|ccc}
\hline
\multirow{2}{*}{Methods} &  \multicolumn{3}{c|}{Lower Jawbone} &  \multicolumn{3}{c|}{Upper Jawbone} &  \multicolumn{3}{c}{Pharynx} \\
\cline{2-10}
& Dice$\uparrow$ & mIoU$\uparrow$ & HD$\downarrow$ & Dice$\uparrow$ & mIoU$\uparrow$ & HD$\downarrow$ & Dice$\uparrow$ & mIoU$\uparrow$ & HD$\downarrow$ \\
\hline
nnUnet \cite{isensee2021nnu}  & 94.55 & 91.37 & 54.64 & 89.28 & 73.98 & 81.31 & 97.00 & 94.32 & 22.96\\
Swin-Umamba \cite{liu2024swin} & 93.59 & 90.31 & 53.07 & 88.92 & 86.99 & 70.64 & 97.01 & 94.34 & 27.30 \\
EMCAD \cite{rahman2024emcad}  & 94.21 & 90.70 & 90.84 & 87.95 & 85.85 & 69.57 & 96.87 & 94.09 & 23.12\\
H-SAM \cite{cheng2024unleashing} & 94.52 & 91.20 & 61.67 & 89.61 & 87.66 & 96.25 & 96.54 & 93.47 & 28.16\\
Zig-RiR \cite{chen2025zig} & 94.26 & 91.05 & 60.76 & 89.70 & 87.71 & 80.07 & 96.83 & 93.81 & 15.33\\
\hline
ToothNet \cite{cui2019toothnet} & 92.73 & 88.96 & 67.89 & 86.54 & 82.71 & 89.42 & 96.28 & 93.58 & 32.71 \\
HMG-Net \cite{cui2021hierarchical} & 93.21 & 89.47 & 63.25 & 87.19 & 83.94 & 85.78 & 96.61 & 93.97 & 31.89 \\
SGA-Net \cite{li2022semantic} & 93.48 & 89.82 & 61.14 & 87.92 & 84.58 & 82.35 & 96.61 & 94.79 & 29.38 \\
T-Mamba \cite{hao2024t} & 94.67 & 91.58 & 48.25 & 90.12 & 88.19 & 65.84 & 97.05 & 94.63 & 21.25 \\
ToothSC-SAM \cite{li2025toothsc} & 94.78 & 91.84 & 45.78 & 90.47 & 88.65 & 62.15 & 97.14 & 94.66 & 19.09 \\
TSG-GCN \cite{liu2024individual} & 94.65 & 91.72 & 47.21 & 90.23 & 88.42 & 64.38 & 97.08 & 94.71 & 20.15 \\
\hline
Co-Seg++  & \textbf{96.54} & \textbf{93.85} & \textbf{38.67} & \textbf{92.78} & \textbf{90.94} & \textbf{52.45} & \textbf{97.73} & \textbf{95.58} & \textbf{13.27}  \\
\hline
\end{tabular}}
\label{tab:sem_tooth}
\end{table*}

\subsection{Optimization Pipeline}
To construct our Co-Seg++ framework, we first adopt Hirea ViT \cite{ryali2023hiera} as the shared image encoder for versatile medical segmentation, ensuring the consistency of feature learning. In particular, we load \cite{ravi2025sam} to initialize corresponding parameters and freeze these weights to preserve pre-trained knowledge. Additionally, we insert Adapter \cite{houlsby2019parameter} into FFN layers to achieve parameter-efficient fine-tuning in medical domains. Based on the co-segmentation paradigm, the predictions for both tasks are based on shared image embeddings and mutual prompt guidance. Specifically, $\theta_i$ and $\theta_j$ contain the parameters of SSP-Encoder and the segmentation head of MTC-Decoder for different tasks. When task $j$ updates its parameters $\theta_j$, it affects the generation of spatial constraints $c_j$ in the first forward, which influence the prediction of task $i$ via the cross-attention mechanism in the second forward. This means that the gradient of task $i$ with respect to task $j$ parameters reflects how shared components allow one task to affect the other. This implicit gradient flow is critical for model performance, as it allows semantic segmentation and instance segmentation to mutually inform each other. In this way, semantic segmentation benefits from instance-level spatial constraints, while instance segmentation predictions are guided by global semantic context.

Moreover, the training of Co-Seg++ consists of two parts: (1) the optimization of probability distribution and prior spatial constraints (2) the joint optimization of semantic and instance segmentation. The overall loss is formulated as:
\begin{equation}
    \mathcal{L}_{\rm CoSeg} = \underbrace{\lambda_1(\mathcal{L}_{\rm SCC} + \mathcal{L}_{\rm BM})}_{\text{Forward 1}} + \underbrace{\mathcal{L}_{\rm SEG}^{\rm sem}+\mathcal{L}_{\rm SEG}^{\rm ins}}_{\text{Forward 2}},
\end{equation}
where $\lambda_1$ are factors to adjust the contribution of the first forward. During backpropagation, the semantic loss in the second forward first influences the gradient updates of the image embeddings, semantic-specific query embeddings and instance-specific prior spatial constraints through the cross-attention mechanism. Specifically, the generation of instance-specific prior spatial constraints depends on binary instance masks that are generated by the image embedding and instance-specific query embeddings in the first forward. This creates a gradient flow pathway where the semantic loss propagates, through the instance-specific prior spatial constraints $c_j$, back to the instance query embeddings that generated the binary masks $y_j^{\rm bin}$ in the first forward. Therefore, the semantic loss directly influences the gradient updates of the instance-specific query embeddings, and vice versa, establishing the bidirectional optimization as described in our co-segmentation paradigm. By optimizing $\mathcal{L}_{\rm CoSeg}$, our framework enables accurate and consistent semantic and instance segmentation through joint learning with spatio-temporal priors and cross-task guidance. Co-Seg++ maintains computational efficiency by leveraging a shared encoder, reusing features across tasks, and introducing only lightweight adapter modules for medical adaptation. This collaborative design not only avoids redundant computation but also allows the two tasks to mutually reinforce each other, leading to more robust predictions. We now proceed to demonstrate the effectiveness of Co-Seg++ through comprehensive experiments across diverse datasets.

\begin{table*}[!t]
% \hspace{-0.6cm}
\centering
\small
\setlength\tabcolsep{3pt}
\caption{Comparison with state-of-the-arts on histopathology instance segmentation.}
\scalebox{0.93}{\begin{tabular}{l|cc|cc|cc|cc|cc|cc|cc|cc|cc}
\hline
& \multicolumn{6}{c|}{PUMA} & \multicolumn{6}{c|}{CRAG} & \multicolumn{6}{c}{GlaS} \\
\cline{2-19}
Methods & \multicolumn{2}{c|}{Tumor} &  \multicolumn{2}{c|}{Lymph} & \multicolumn{2}{c|}{Other Cells} & \multicolumn{2}{c|}{Epithelial} &  \multicolumn{2}{c|}{LR} &  \multicolumn{2}{c|}{Connective} & \multicolumn{2}{c|}{Epithelial} &  \multicolumn{2}{c|}{LR} &  \multicolumn{2}{c}{Connective} \\
\cline{2-19}
& F1$\uparrow$ & AJI$\uparrow$ & F1$\uparrow$ & AJI$\uparrow$ & F1$\uparrow$ & AJI$\uparrow$ & F1$\uparrow$ & AJI$\uparrow$ & F1$\uparrow$ & AJI$\uparrow$ & F1$\uparrow$ & AJI$\uparrow$ & F1$\uparrow$ & AJI$\uparrow$ & F1$\uparrow$ & AJI$\uparrow$ & F1$\uparrow$ & AJI$\uparrow$ \\
\hline
nnUnet \cite{isensee2021nnu}  & 65.84 & 52.17 & 54.38 & 33.62 & 19.76 & 12.43 & 58.92 & 46.25 & 43.85 & 31.74 & 56.73 & 45.82 & 46.52 & 32.89 & 55.38 & 35.62 & 24.17 & 15.83 \\
Swin-Umamba \cite{liu2024swin}  & 68.73 & 56.42 & 57.29 & 36.85 & 22.64 & 14.76 & 61.48 & 49.33 & 46.52 & 34.17 & 59.35 & 48.27 & 49.16 & 35.44 & 57.93 & 38.29 & 26.75 & 17.92 \\
EMCAD \cite{rahman2024emcad}  & 70.15 & 58.93 & 58.74 & 38.42 & 24.83 & 16.29 & 63.27 & 51.08 & 48.16 & 35.73 & 61.29 & 50.14 & 51.38 & 37.62 & 59.74 & 40.16 & 28.49 & 19.35 \\
H-SAM \cite{cheng2024unleashing}  & 74.32 & 62.58 & 61.47 & 40.93 & 28.56 & 18.74 & 67.85 & 55.29 & 52.84 & 39.16 & 65.73 & 54.37 & 56.29 & 41.85 & 62.47 & 42.93 & 33.18 & 22.67 \\
Zig-RiR \cite{chen2025zig} & 71.89 & 60.24 & 59.83 & 39.57 & 26.47 & 17.35 & 65.18 & 52.76 & 50.32 & 37.48 & 63.41 & 52.15 & 53.74 & 39.17 & 60.83 & 41.52 & 30.64 & 20.81 \\
\hline
HoverNet \cite{graham2019hover}& 74.86 & 63.42 & 62.35 & 41.18 & 28.67 & 19.15 & 68.73 & 55.94 & 53.81 & 40.26 & 66.94 & 55.47 & 57.92 & 42.74 & 64.29 & 43.85 & 34.18 & 23.59 \\
HookNet \cite{van2021hooknet} & 72.61 & 60.63 & 61.21 & 39.80 & 24.12 & 15.28 & 64.74 & 52.16 & 49.79 & 36.29 & 62.27 & 50.98 & 51.69 & 37.33 & 61.16 & 39.77 & 29.83 & 19.68\\
CellPose \cite{stringer2021cellpose}  & 75.34 & 63.91 & 63.17 & 41.73 & 29.45 & 19.82 & 69.28 & 56.63 & 54.15 & 40.67 & 67.41 & 55.83 & 58.26 & 43.12 & 64.73 & 44.18 & 35.47 & 24.26 \\
CPPNet \cite{chen2023cpp} & 73.28 & 61.74 & 62.15 & 40.67 & 27.89 & 18.43 & 66.47 & 53.82 & 51.63 & 38.27 & 64.35 & 52.78 & 54.86 & 40.25 & 62.73 & 42.68 & 32.17 & 21.94 \\
CellViT \cite{horst2024cellvit} & 75.19 & 63.85 & 62.94 & 41.57 & 29.73 & 19.68 & 69.16 & 56.42 & 54.29 & 40.83 & 67.58 & 55.94 & 62.47 & 45.16 & 64.85 & 44.27 & 35.62 & 24.38 \\
PromptNucSeg \cite{shui2024unleashing} & 74.63 & 63.17 & 61.98 & 40.85 & 28.94 & 19.26 & 68.35 & 55.73 & 53.74 & 40.15 & 66.82 & 55.29 & 63.63 & 45.48 & 64.12 & 43.69 & 34.86 & 23.74 \\
PathoSAM \cite{griebel2025segment}  & 76.54 & 65.62 & 60.76 & 41.10 & 31.20 & 20.55 & 73.13 & 58.52 & 67.09 & 50.33 & 70.12 & 58.25 & 63.86 & 46.56 & 63.59 & 41.30 & 40.39 & 27.36  \\
\hline
Co-Seg++ & \textbf{80.25} & \textbf{70.59} & \textbf{64.75} & \textbf{45.03} & \textbf{36.91} & \textbf{23.65} & \textbf{77.06} & \textbf{61.63} & \textbf{69.71} & \textbf{53.66} & \textbf{74.68} & \textbf{61.37} & \textbf{67.81} & \textbf{50.23} & \textbf{69.46} & \textbf{46.53} & \textbf{47.71} & \textbf{33.31} \\
\hline
\end{tabular}}
\label{tab:ins_histo}
\end{table*}

\begin{table}[!t]
% \hspace{-0.6cm}
\centering
\small
\setlength\tabcolsep{9.5pt}
\caption{Comparison with state-of-the-arts on CBCT instance segmentation.}
\scalebox{1}{\begin{tabular}{l|cc|cc}
\hline
\multirow{2}{*}{Methods} &  \multicolumn{2}{c|}{Lower Teeth} &  \multicolumn{2}{c}{Upper Teeth}  \\
\cline{2-5}
& F1$\uparrow$ & AJI$\uparrow$ & F1$\uparrow$ & AJI$\uparrow$ \\
\hline
nnUnet \cite{isensee2021nnu}  & 74.28 & 68.15 & 62.74 & 56.83 \\
Swin-Umamba \cite{liu2024swin}  & 72.85 & 66.42 & 60.37 & 54.29 \\
EMCAD \cite{rahman2024emcad}  & 73.96 & 67.58 & 61.84 & 55.67 \\
H-SAM \cite{cheng2024unleashing}  & 75.43 & 69.15 & 63.28 & 57.15 \\
Zig-RiR \cite{chen2025zig} & 74.67 & 68.29 & 62.55 & 56.41 \\
\hline
ToothNet \cite{cui2019toothnet}& 77.16 & 71.05 & 64.30 & 58.32 \\
HMG-Net \cite{cui2021hierarchical}  & 76.83 & 70.67 & 65.12 & 59.24 \\
SGA-Net \cite{li2022semantic} & 77.94 & 72.36 & 66.47 & 60.85 \\
T-Mamba \cite{hao2024t} & 78.26 & 72.71 & 67.15 & 61.38 \\
ToothSC-SAM \cite{li2025toothsc} & 79.15 & 73.42 & 68.29 & 62.74 \\
TSG-GCN \cite{liu2024individual} & 78.56 & 71.41 & 69.18 & 62.48\\
\hline
Co-Seg++  & \textbf{81.99} & \textbf{74.15} & \textbf{72.41} & \textbf{66.92} \\
\hline
\end{tabular}}
\label{tab:ins_tooth}
\end{table}

\section{Experiments}
\subsection{Datasets and Implementations}
\subsubsection{Datasets}
To validate the effectiveness of the proposed Co-Seg++, we conduct comprehensive evaluations across diverse medical domains, including three H\&E-stained histopathology datasets: PUMA \cite{giaf011}, GlaS \cite{sirinukunwattana2017gland}, CRAG \cite{graham2023one} and one CBCT dataset: ToothFairy2 \cite{bolelli2024segmenting}, as follows:

\noindent \textbf{PUMA} \cite{giaf011} is a melanoma nuclei and tissue segmentation dataset containing 206 pathological slides scanned at 40$\times$ magnification with a resolution of $1024 \times 1024$. The dataset covers two tissue regions (\textit{i.e.}, tumor and stroma) and three nuclei types (\textit{i.e.}, tumor, lymphocyte, and other cells). 

\noindent \textbf{GlaS} \cite{sirinukunwattana2017gland} dataset, derived from the 2015 Gland Segmentation Challenge, comprises 165 histopathology images captured at 20$\times$ magnification with glandular regions and epithelial, connective and lympho-reticular nuclei instances. The image size varies from $574 \times 433$ to $775 \times 522$.

\noindent \textbf{CRAG} \cite{graham2023one} dataset contains 216 colorectal adenocarcinoma patches at 20x magnification with a resolution of $996 \times 996$. The dataset annotates glandular regions and epithelial, connective and lympho-reticular nuclei instances.

\noindent \textbf{ToothFairy2} \cite{bolelli2024segmenting} comprises 480 cases with comprehensive annotations for dental anatomical structures, including jawbone and pharynx regions and tooth instances. The image size varies from $272 \times 345$ to $512 \times 512$.

\subsubsection{Implementation Details}
We perform all experiments on a workstation equipped with an Intel Xeon Silver 4216 CPU, 256GB RAM, and an NVIDIA A5000 GPU (24GB memory). The software environment consists of PyTorch 2.4 and CUDA 12.8. For fair comparisons, all semantic and instance segmentation methods are implemented with the same training settings and configurations. We utilize the pre-trained SAM ViT-H \cite{kirillov2023_sam} structure as the image encoder of medical SAM baselines. We adopt the Hirea-L structure \cite{ryali2023hiera} as the image encoder of our Co-Seg++. We apply the optimizer using Adam with an initial learning rate of $1\times10^{-4}$ and use the exponential decay strategy to adjust the learning rate with a factor of 0.98. The loss coefficient $\lambda_1$ is set to 0.5. We split all datasets into the training, validation, and test sets as 7:1:2, and all images are resized to $1024 \times 1024$ during the fine-tuning and evaluation stage \cite{ma2024segment}. The batch size and the training epoch are set to 16 and 300. Our Co-Seg++ introduces 8.3\% additional learnable parameters compared to the baseline in Table \ref{tab:ablation}.

\subsubsection{Evaluation Metrics}
To perform a comprehensive evaluation of versatile medical segmentation, we apply different metrics in terms of three segmentation tasks. For semantic segmentation, we select the Dice coefficient, mean intersection over union (mIoU), and Hausdorff distance (HD). For instance segmentation, we compare the performance with four metrics, including object-level F1 score and aggregated Jaccard index (AJI). For panoptic segmentation, we adopt panoptic quality (PQ). Except for HD, which measures the distance between predicted and ground truth boundary points, higher scores for these metrics indicate better segmentation quality.

\begin{table*}[!t]
\centering
\small
\setlength\tabcolsep{1.5pt}
\caption{Comparison with state-of-the-arts on histopathology panoptic segmentation.}
\scalebox{1}{\begin{tabular}{l|ccccc|cccc|cccc}
\hline
& \multicolumn{5}{c|}{PUMA} & \multicolumn{4}{c|}{CRAG} & \multicolumn{4}{c}{GlaS} \\
\cline{2-14}
Methods & \multicolumn{2}{c|}{Semantic} & \multicolumn{3}{c|}{Instance} & \multicolumn{1}{c|}{Semantic} & \multicolumn{3}{c|}{Instance} & \multicolumn{1}{c|}{Semantic} & \multicolumn{3}{c}{Instance} \\
\cline{2-14}
& Tumor & \multicolumn{1}{c|}{Stroma} & Tumor & Lymph & Other Cells & \multicolumn{1}{c|}{Gland} & Epithelial & LR & Connective & \multicolumn{1}{c|}{Gland} & Epithelial & LR & Connective \\
\hline
nnUnet \cite{isensee2021nnu} & 57.84 & \multicolumn{1}{c|}{35.72} & 59.47 & 49.31 & 22.64 & \multicolumn{1}{c|}{64.26} & 52.38 & 48.79 & 51.27 & \multicolumn{1}{c|}{55.39} & 44.28 & 46.94 & 29.03 \\
Swin-Umamba \cite{liu2024swin} & 58.96 & \multicolumn{1}{c|}{36.45} & 60.73 & 50.68 & 23.17 & \multicolumn{1}{c|}{65.84} & 53.71 & 49.82 & 52.59 & \multicolumn{1}{c|}{56.85} & 45.13 & 47.86 & 29.67 \\
EMCAD \cite{rahman2024emcad} & 59.42 & \multicolumn{1}{c|}{36.89} & 61.15 & 51.04 & 23.85 & \multicolumn{1}{c|}{66.38} & 54.29 & 50.34 & 53.18 & \multicolumn{1}{c|}{57.48} & 45.72 & 48.41 & 30.15 \\
H-SAM \cite{cheng2024unleashing} & 60.73 & \multicolumn{1}{c|}{37.52} & 62.58 & 51.94 & 25.94 & \multicolumn{1}{c|}{67.29} & 55.18 & 51.26 & 53.97 & \multicolumn{1}{c|}{58.16} & 46.42 & 49.28 & 30.84 \\
Zig-RiR \cite{chen2025zig} & 60.15 & \multicolumn{1}{c|}{37.18} & 62.07 & 51.63 & 25.31 & \multicolumn{1}{c|}{66.82} & 54.75 & 50.89 & 53.54 & \multicolumn{1}{c|}{57.73} & 46.09 & 48.95 & 30.46 \\
\hline
HoverNet \cite{graham2019hover} & 60.85 & \multicolumn{1}{c|}{37.68} & 62.74 & 52.15 & 26.37 & \multicolumn{1}{c|}{67.52} & 55.47 & 51.63 & 54.29 & \multicolumn{1}{c|}{58.47} & 46.71 & 49.58 & 30.96 \\
HookNet \cite{van2021hooknet} & 58.73 & \multicolumn{1}{c|}{36.21} & 60.38 & 50.42 & 24.72 & \multicolumn{1}{c|}{65.29} & 53.18 & 49.35 & 52.14 & \multicolumn{1}{c|}{56.28} & 44.89 & 47.53 & 29.48 \\
CellPose \cite{stringer2021cellpose} & 61.29 & \multicolumn{1}{c|}{37.94} & 63.15 & 52.48 & 26.85 & \multicolumn{1}{c|}{68.14} & 55.83 & 51.17 & 54.72 & \multicolumn{1}{c|}{58.96} & 47.15 & 50.14 & 31.18 \\
CPPNet \cite{chen2023cpp} & 60.74 & \multicolumn{1}{c|}{37.16} & 62.47 & 51.69 & 26.34 & \multicolumn{1}{c|}{67.45} & 55.24 & 52.48 & 53.16 & \multicolumn{1}{c|}{60.38} & 45.52 & 48.47 & 30.52 \\
CellViT \cite{horst2024cellvit} & 61.95 & \multicolumn{1}{c|}{38.28} & 64.72 & 52.81 & 27.69 & \multicolumn{1}{c|}{68.73} & 56.49 & 52.69 & 55.41 & \multicolumn{1}{c|}{61.67} & 47.73 & 50.68 & 31.74 \\
PromptNucSeg \cite{shui2024unleashing} & 61.83 & \multicolumn{1}{c|}{38.19} & 63.58 & 52.74 & 27.52 & \multicolumn{1}{c|}{68.56} & 56.38 & 52.59 & 55.28 & \multicolumn{1}{c|}{62.52} & 47.64 & 50.59 & 31.65 \\
PathoSAM \cite{griebel2025segment} & 62.17 & \multicolumn{1}{c|}{38.45} & 65.01 & 52.86 & 27.88 & \multicolumn{1}{c|}{69.98} & 56.22 & 52.45 & 56.05 & \multicolumn{1}{c|}{62.42} & 47.54 & 50.42 & 31.20 \\
\hline
Co-Seg++ & \textbf{64.52} & \multicolumn{1}{c|}{\textbf{41.67}} & \textbf{67.37} & \textbf{54.60} & \textbf{29.48} & \multicolumn{1}{c|}{\textbf{72.69}} & \textbf{59.61} & \textbf{54.90} & \textbf{58.84} & \multicolumn{1}{c|}{\textbf{65.32}} & \textbf{49.01} & \textbf{53.04} & \textbf{34.24} \\
\hline
\end{tabular}}
\label{tab:pano_histo}
\end{table*}

\begin{table}[!t]
% \hspace{-0.6cm}
\centering
\small
\setlength\tabcolsep{3pt}
\caption{Comparison with state-of-the-arts on CBCT panoptic segmentation.}
\scalebox{1}{\begin{tabular}{l|cc|cc|c}
\hline
\multirow{2}{*}{Methods} &  \multicolumn{2}{c|}{Lower} &  \multicolumn{2}{c|}{Upper} 
& \multirow{2}{*}{Pharynx}\\
\cline{2-5}
& Jawbone & Teeth & Jawbone & Teeth & \\
\hline
nnUnet \cite{isensee2021nnu}  & 75.82 & 61.46 & 48.73 & 52.35 & 84.27 \\
Swin-Umamba \cite{liu2024swin}  & 76.94 & 62.85 & 49.58 & 53.72 & 85.69 \\
EMCAD \cite{rahman2024emcad}  & 77.63 & 63.49 & 50.21 & 54.38 & 86.45 \\
H-SAM \cite{cheng2024unleashing}  & 78.85 & 64.92 & 51.47 & 55.83 & 87.94 \\
Zig-RiR \cite{chen2025zig} & 78.32 & 64.17 & 50.89 & 55.24 & 87.38 \\
VerSemi \cite{zeng2025segment} & 80.16 & 65.74 & 52.38 & 57.52 & 89.88 \\
\hline
ToothNet \cite{cui2019toothnet}& 77.47 & 62.73 & 48.14 & 52.47 & 85.62 \\
HMG-Net \cite{cui2019toothnet}  & 78.18 & 64.29 & 49.86 & 54.12 & 86.25 \\
SGA-Net \cite{li2022semantic} & 80.24 & 66.58 & 52.74 & 57.29 & 89.43 \\
T-Mamba \cite{hao2024t} & 79.67 & 66.15 & 52.18 & 56.81 & 88.87 \\
ToothSC-SAM \cite{li2025toothsc} & 80.93 & 67.42 & 53.04 & 57.65 & 90.16 \\
TSG-GCN \cite{liu2024individual} & 81.00 & 67.74 & 53.26 & 57.98 & 90.31 \\
\hline
Co-Seg++  & \textbf{83.25} & \textbf{70.84} & \textbf{57.57} & \textbf{61.66} & \textbf{94.49}\\
\hline
\end{tabular}}
\label{tab:pano_tooth}
\end{table}

\subsection{Comparison on Medical Semantic Segmentation}
To comprehensively evaluate the performance of Co-Seg++, we first perform comparisons with state-of-the-art methods on medical semantic segmentation across three histopathology datasets. As shown in Table \ref{tab:sem_histo}, classic semantic segmentation methods \cite{isensee2021nnu, rahman2024emcad} generally underperform compared to prompt-based frameworks \cite{cheng2024unleashing}. On the PUMA dataset, H-SAM \cite{cheng2024unleashing} achieves competitive results with 91.31\% Dice for tumor segmentation, while specialized histopathology methods such as PromptNucSeg \cite{shui2024unleashing} and PathoSAM \cite{griebel2025segment} demonstrate superior performance with 91.57\% and 91.78\% Dice, respectively. In contrast, our Co-Seg++ substantially outperforms all baseline methods across all datasets. More notably, for the challenging stroma segmentation task, Co-Seg++ demonstrates significant superiority with 60.88\% Dice, representing a substantial 5.54\% improvement over the best baseline Zig-RiR \cite{chen2025zig}.

Table \ref{tab:sem_tooth} demonstrates the effectiveness of Co-Seg++ on dental CBCT semantic segmentation. We observe that general medical segmentation methods (e.g, nnUNet \cite{isensee2021nnu}) achieve competitive baseline performance, while recent Mamba and prompt approaches further promote dental segmentation. Remarkably, Co-Seg++ achieves state-of-the-art performance across all anatomical structures with statistical significance (\textit{e.g.}, P-value $<$ 0.005). For lower jawbone segmentation, Co-Seg++ reveals 96.54\% Dice and 93.85\% mIoU, representing a 1.76\% Dice improvement over the best baseline ToothSC-SAM \cite{li2025toothsc}. The improvement is even more pronounced for upper jawbone segmentation, where Co-Seg++ achieves 92.78\% Dice with a 2.31\% improvement. For pharynx segmentation, Co-Seg++ demonstrates consistent superiority with 97.73\% Dice and the lowest HD of 13.27, indicating exceptional precision in boundary delineation. These comparisons validate the superior performance of our Co-Seg++ on diverse medical semantic segmentation tasks through mutual optimization.

\subsection{Comparison on Medical Instance Segmentation}

We further evaluate Co-Seg++ on medical instance segmentation tasks to demonstrate the effectiveness of our collaborative learning framework in handling complex instance-level segmentation challenges. Table \ref{tab:ins_histo} presents comprehensive results on histopathology instance segmentation across multiple cell types and datasets. We observe that specialized histopathology methods generally outperform generic segmentation approaches. CellPose \cite{horst2024cellvit} and PathoSAM \cite{griebel2025segment} demonstrate superior performance compared to traditional methods such as nnUNet \cite{isensee2021nnu}. Our Co-Seg++ consistently surpasses all baselines by substantial margins across all datasets and cell categories with a P-value $< 0.001$. On the PUMA dataset, Co-Seg++ achieves remarkable performance for tumor instance segmentation with 80.25\% F1 score and 70.59\% AJI, representing significant improvements of 3.71\% F1 and 4.97\% AJI over the best baseline PathoSAM \cite{griebel2025segment}.

As shown in Table \ref{tab:ins_tooth}, Co-Seg++ demonstrates superior performance on dental CBCT instance segmentation tasks. We observe progressive improvements from general medical methods to specialized dental segmentation approaches, with ToothSC-SAM \cite{li2025toothsc} and T-Mamba \cite{hao2024t} representing strong baselines. Co-Seg++ achieves exceptional results with 81.99\% F1 score for lower teeth segmentation, representing improvements of 2.84\% F1 score over the best baseline ToothSC-SAM \cite{li2025toothsc}. For upper teeth segmentation, which is typically more challenging due to anatomical complexity, Co-Seg++ demonstrates substantial superiority with 72.41\% F1 and 66.92\% AJI, achieving improvements of 3.23\% F1 and 4.44\% AJI, respectively. These results prove the effectiveness of our cross-guidance approach in handling complex instance boundaries and overlapping structures common in medical imaging.

\begin{figure*}[!t]
  % \vspace{-4.3cm}
  \centering
  \includegraphics[width=0.95\linewidth]{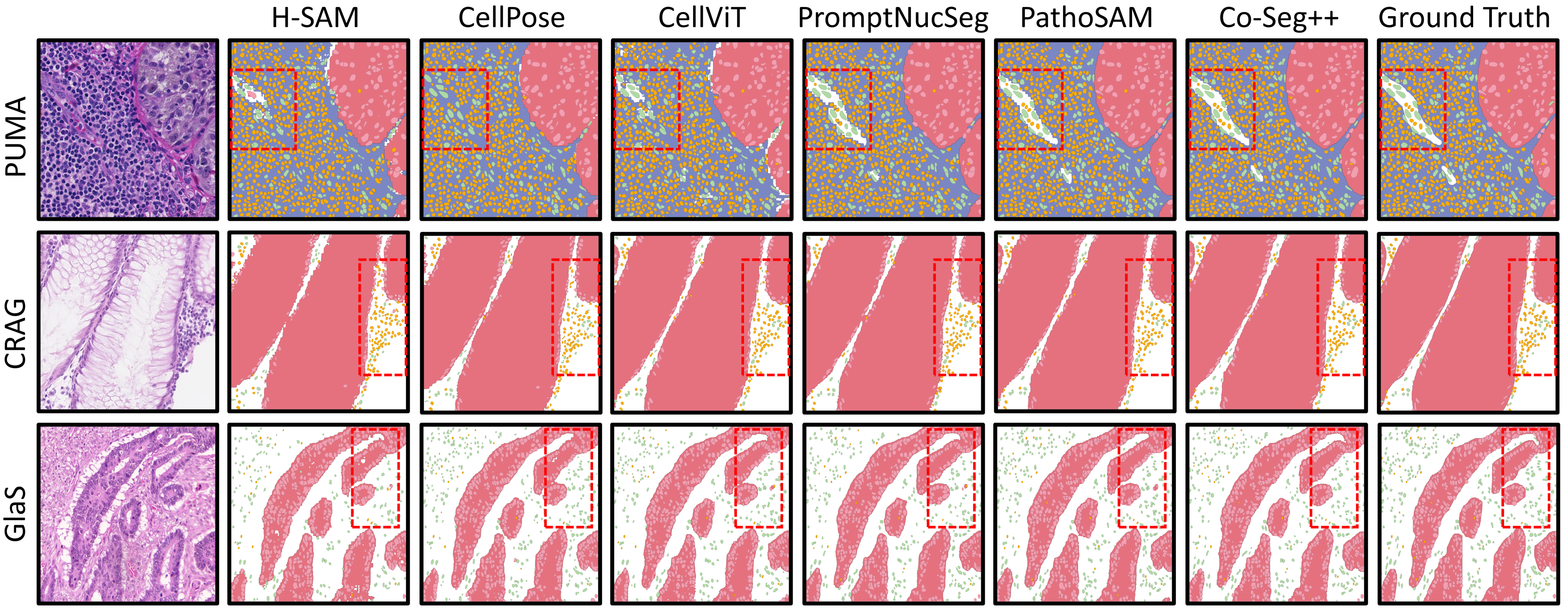}
  \caption{Visualization of tissue\/gland semantic and nuclei instance segmentation. Our Co-Seg++ exhibits the best results, recognizing more nuclei instances with accurate categories and tissue\/gland boundaries while having fewer false positives.}
  \label{fig:visual}
  % \vspace{-1.0em}
\end{figure*}

\begin{table*}[!t]
\centering
\small
\setlength\tabcolsep{2.5pt}
\caption{Ablation Study of Co-Seg++ on Histopathology Panoptic Segmentation. }
\scalebox{1}{\begin{tabular}{cccc|ccccc|cccc|cccc}
\hline
& \multirow{3}{*}{$P$} & \multirow{3}{*}{$D$} & \multirow{3}{*}{$C$} &\multicolumn{5}{c|}{PUMA} & \multicolumn{4}{c|}{CRAG} & \multicolumn{4}{c}{GlaS} \\
\cline{5-17}
& & & & \multicolumn{2}{c|}{Semantic} & \multicolumn{3}{c|}{Instance} & \multicolumn{1}{c|}{Semantic} & \multicolumn{3}{c|}{Instance} & \multicolumn{1}{c|}{Semantic} & \multicolumn{3}{c}{Instance} \\
\cline{5-17}
& & & & Tumor & \multicolumn{1}{c|}{Stroma} & Tumor & Lymph & Other Cells & \multicolumn{1}{c|}{Gland} & Epithelial & LR & Connective & \multicolumn{1}{c|}{Gland} & Epithelial & LR & Connective \\
\hline
1 &  &  &  & 61.85 & \multicolumn{1}{c|}{37.92} & 64.58 & 52.41 & 27.35 & \multicolumn{1}{c|}{68.45} & 55.72 & 51.88 & 55.16 & \multicolumn{1}{c|}{61.94} & 46.85 & 49.74 & 31.08 \\
2 & \checkmark & & & 62.78 & \multicolumn{1}{c|}{38.65} & 65.89 & 53.15 & 28.12 & \multicolumn{1}{c|}{69.87} & 56.89 & 52.94 & 56.28 & \multicolumn{1}{c|}{63.18} & 47.69 & 50.85 & 32.05 \\
3 & & \checkmark & & 62.34 & \multicolumn{1}{c|}{38.28} & 65.12 & 52.73 & 27.68 & \multicolumn{1}{c|}{69.02} & 56.14 & 52.25 & 55.63 & \multicolumn{1}{c|}{62.47} & 47.21 & 50.18 & 31.46 \\
4 & & & \checkmark & 63.24 & \multicolumn{1}{c|}{39.18} & 66.45 & 53.52 & 28.59 & \multicolumn{1}{c|}{70.52} & 57.48 & 53.67 & 57.01 & \multicolumn{1}{c|}{63.89} & 48.15 & 51.48 & 32.68 \\
5 & \checkmark & \checkmark &  & 63.15 & \multicolumn{1}{c|}{39.02} & 66.28 & 53.38 & 28.47 & \multicolumn{1}{c|}{70.31} & 57.29 & 53.44 & 56.79 & \multicolumn{1}{c|}{63.72} & 48.02 & 51.29 & 32.52 \\
6 & \checkmark & & \checkmark & 63.82 & \multicolumn{1}{c|}{40.12} & 66.91 & 53.99 & 29.08 & \multicolumn{1}{c|}{71.84} & 58.47 & 54.32 & 57.92 & \multicolumn{1}{c|}{64.55} & 48.69 & 52.18 & 33.31 \\
7 & & \checkmark & \checkmark & 63.58 & \multicolumn{1}{c|}{39.54} & 66.82 & 53.89 & 28.92 & \multicolumn{1}{c|}{71.18} & 57.92 & 54.21 & 57.45 & \multicolumn{1}{c|}{64.41} & 48.58 & 51.96 & 33.15 \\
8 & \checkmark & \checkmark & \checkmark & \textbf{64.52} & \multicolumn{1}{c|}{\textbf{41.67}} & \textbf{67.37} & \textbf{54.60} & \textbf{29.48} & \multicolumn{1}{c|}{\textbf{72.69}} & \textbf{59.61} & \textbf{54.90} & \textbf{58.84} & \multicolumn{1}{c|}{\textbf{65.32}} & \textbf{49.01} & \textbf{53.04} & \textbf{34.24}  \\
\hline
\end{tabular}}
\label{tab:ablation}
\end{table*}

\begin{figure}[!t]
  % \vspace{-4.3cm}
  \centering
  \includegraphics[width=0.98\linewidth]{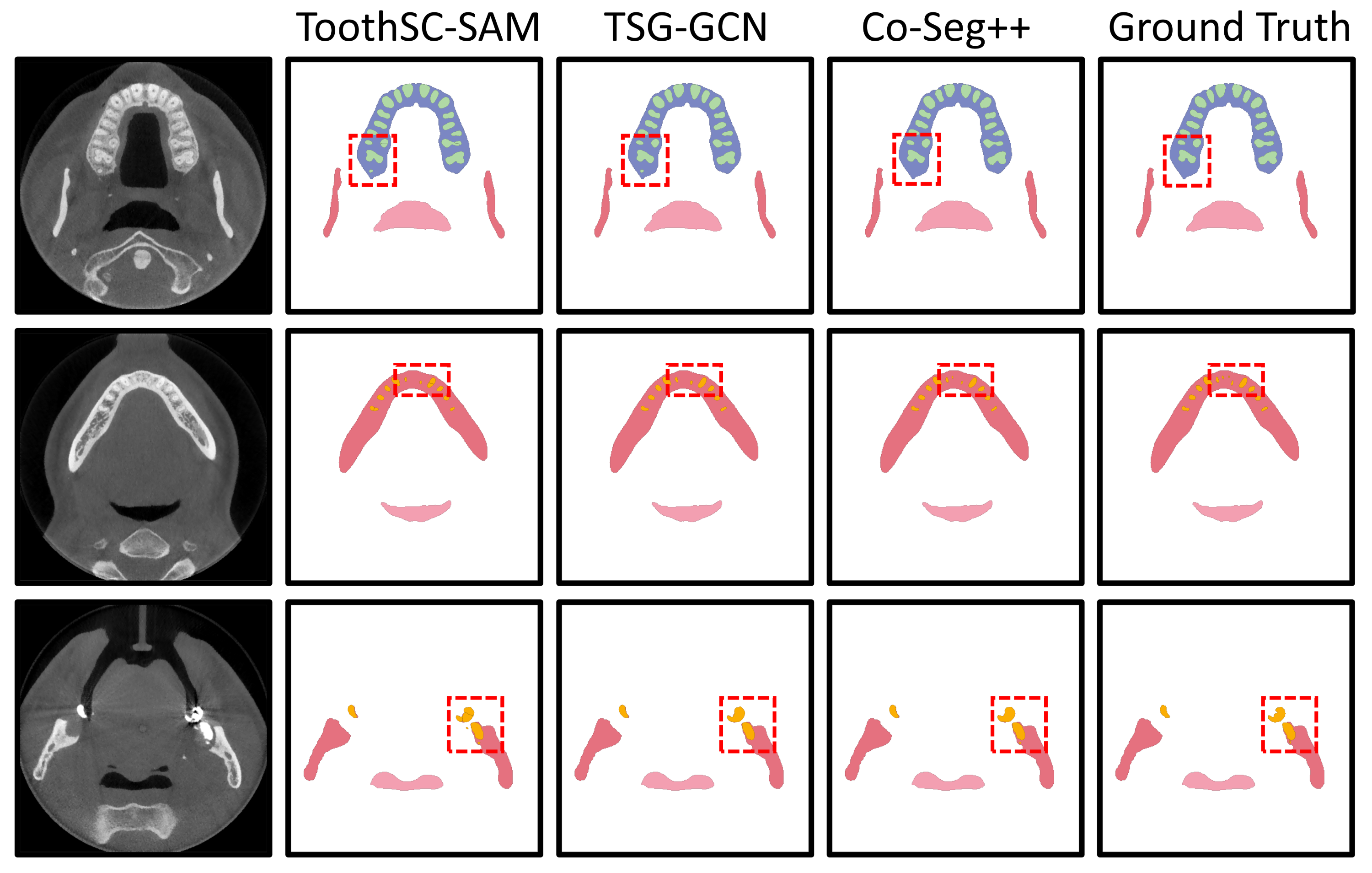}
  \caption{Visualization of CBCT semantic and instance segmentation. Benefiting from the mutual guidance of both segmentation tasks, Co-Seg++ delineates precise jawbone areas and segments accurate teeth.}
  \label{fig:visual_teeth}
  % \vspace{-1.0em}
\end{figure}

\subsection{Comparison on Medical Panoptic Segmentation}

To comprehensively evaluate the versatility of Co-Seg++, we conduct extensive experiments on medical panoptic segmentation, which simultaneously requires both semantic understanding and instance-level discrimination capabilities. Table \ref{tab:pano_histo} demonstrates the superior performance of Co-Seg++ on histopathology panoptic segmentation across diverse tissue, gland and nuclei types. We observe that specialized methods like CellViT \cite{horst2024cellvit} and PathoSAM \cite{griebel2025segment} achieve competitive performance compared to general segmentation approaches, with PathoSAM \cite{griebel2025segment} showing strong results across most categories. In contrast, Co-Seg++ consistently outperforms all baseline methods with substantial improvements and statistical significance (\textit{e.g.}, with a P-value $< 0.001$) across all anatomical structures and datasets. On the PUMA dataset, Co-Seg++ achieves remarkable performance for tumor panoptic quality with 64.52\% (2.35\% improvement over PathoSAM \cite{griebel2025segment}) and demonstrates particularly significant improvement for the challenging stroma segmentation task with 41.67\% (3.22\% improvement). Moreover, we compare the quantitative results of our Co-Seg++ with the best baseline methods in Fig.~\ref{fig:visual} and our Co-Seg++ can segment nuclei more accurately with better tissue and gland regions.

We further perform comparisons on dental CBCT panoptic segmentation, as shown in Table \ref{tab:pano_tooth}. SGA-Net \cite{li2022semantic}, ToothSC-SAM \cite{li2025toothsc}, and TSG-GCN \cite{liu2024individual} achieve competitive results, with TSG-GCN showing particularly strong performance across multiple anatomical structures. Co-Seg++ achieves state-of-the-art performance across all dental anatomical components, by a remarkable PQ increase of 3.1\% and 3.68\% on lower and upper teeth segmentation, respectively. We also elaborate on the panoptic segmentation results in Fig~\ref{fig:visual_teeth}. The comprehensive evaluation across panoptic segmentation tasks validates the effectiveness of our mutual prompt-guided collaborative learning in handling the complex interplay between semantic regions and instance boundaries.

\subsection{Ablation Study}
To investigate the effectiveness of the SSP-Encoder $P$, MTC-Decoder $D$, and co-segmentation paradigm $C$, we conduct comprehensive ablation studies on histopathology panoptic segmentation across three datasets (PUMA, CRAG, and GlaS), as illustrated in Table \ref{tab:ablation}. By removing the tailored modules from Co-Seg, we construct an original encoder-shared framework with Hiera-L \cite{ravi2025sam} as the ablation baseline. By separately introducing the SSP-Encoder ($2^{nd}$ row), MTC-Decoder ($3^{rd}$ row), and co-segmentation paradigm ($4^{th}$ row), the performance demonstrates consistent improvements across all datasets. Notably, our co-segmentation paradigm achieves the most significant individual contribution, with substantial improvements averaging PQ of 1.61\%. We additionally investigate the effects of combined components. By comparing $6^{th}$ and $7^{th}$ rows with the $5^{th}$ row, configurations with co-segmentation demonstrate significant gains, achieving an average PQ of 53.48\% and 53.20\%, respectively. On this basis, our complete Co-Seg++ framework ($8^{th}$ row) simultaneously adopts all three components to achieve the best performance across all datasets and categories. The full model attains optimal performance with an average PQ of 54.25\%, representing an average improvement of 3.10\% over the baseline. These comprehensive ablation experiments validate that the tailored SSP-Encoder, MTC-Decoder, and co-segmentation paradigm collectively contribute to the superior performance of Co-Seg++ by effectively capturing long-range dependencies, enabling cross-task collaboration, and facilitating mutual enhancement between semantic and instance segmentation.

Furthermore, we investigate the effect of SiLU activation and gating operations (GO) on our Co-Seg++, as illustrated in Table \ref{tab:ablation_ssp}. By introducing the SiLU activation function, we observe performance degradation across all categories, with semantic segmentation dropping from 64.52\% to 60.17\% for tumor regions and from 41.67\% to 36.73\% for stroma regions. Similarly, instance segmentation performance decreases from 67.37\% to 62.15\% for tumor nuclei and from 54.60\% to 51.71\% for lymphocytes. When both SiLU and gating operations are added, the performance shows slight improvement compared to SiLU-only but still remains inferior to our streamlined design. Therefore, the additional complexity introduced by SiLU activation and gating operations is not beneficial for our framework. These results demonstrate that the design of our SSP-Encoder is both effective and efficient, achieving superior performance in versatile medical segmentation.

\begin{table}[!t]
\centering
\small
\setlength\tabcolsep{2.3pt}
\caption{Ablation Study of SSP-Encoder on PUMA panoptic segmentation.}
\scalebox{0.91}{\begin{tabular}{l|cc|ccc}
\hline
\multirow{2}{*}{Methods} & \multicolumn{2}{c|}{Semantic} & \multicolumn{3}{c}{Instance}\\
\cline{2-6}
& Tumor & Stroma & Tumor & Lymph & Other Cells\\
\hline
SSP-Encoder + SiLU & 60.17 & 36.73 & 62.15 & 51.71 & 25.46 \\
SSP-Encoder + SiLU + GO & 62.03 & 38.09 & 64.92 & 52.43 & 27.75 \\
\hline
SSP-Encoder (Ours) & \textbf{64.52} & \textbf{41.67} & \textbf{67.37} & \textbf{54.60} & \textbf{29.48} \\

\hline
\end{tabular}}
\label{tab:ablation_ssp}
\end{table}

\begin{table}[!t]
\centering
\caption{Comparison of computation cost on histopathology semantic and instance segmentation.}
\setlength\tabcolsep{2pt}
\label{tab:efficiency}
\scalebox{0.92}{\begin{tabular}{l|cccc|cc}
\hline
Methods & FLOPs (G)$\downarrow$ & \begin{tabular}[c]{@{}c@{}}Inference\\Latency (ms)$\downarrow$\end{tabular} & \begin{tabular}[c]{@{}c@{}}Params\\(M)$\downarrow$\end{tabular} & \begin{tabular}[c]{@{}c@{}}Learnable\\Params (M)$\downarrow$\end{tabular} & Dice$\uparrow$ & AJI$\uparrow$ \\
\hline
CellViT \cite{horst2024cellvit} & 4896.15 & 823.68 & 891.29 & 891.29 & 80.61 & 43.57 \\
PathoSAM \cite{griebel2025segment} & 2739.59 & 448.33 & 645.16 & 645.16 & 81.46 & 45.53 \\
\hline
Co-Seg++ & \textbf{869.67} & \textbf{186.53} & \textbf{234.33} & \textbf{22.18} & \textbf{84.09} & \textbf{49.56} \\
\hline
\end{tabular}}
\end{table}

\begin{figure}[!t]
  % \vspace{-4.3cm}
  \centering
  \includegraphics[width=1\linewidth]{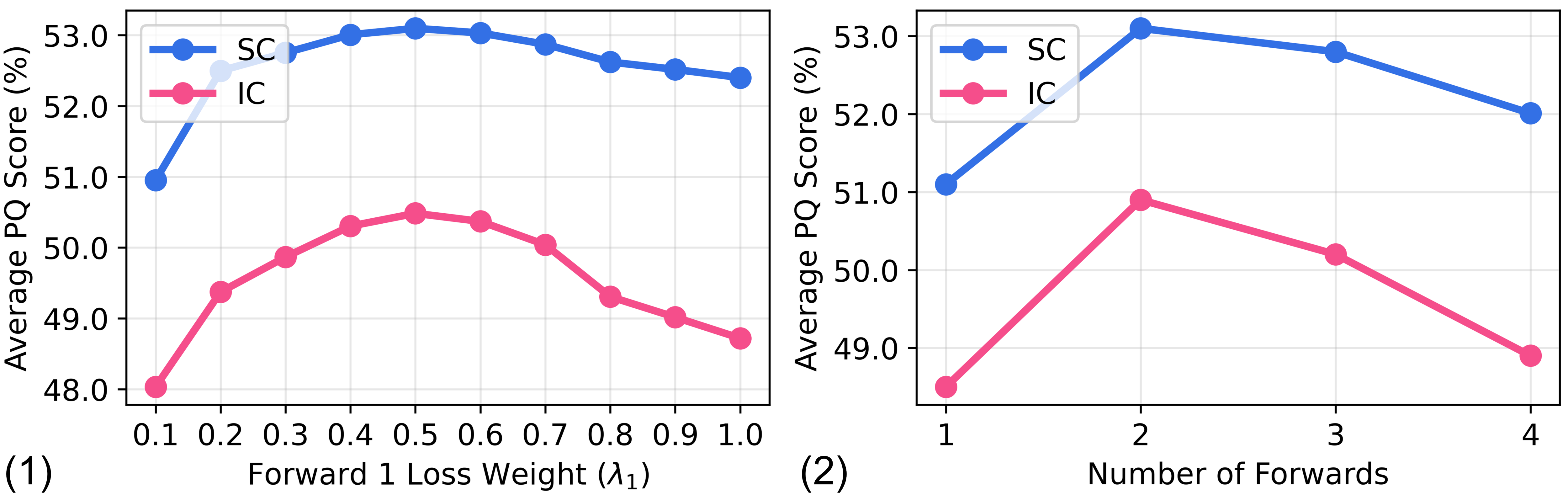}
  \caption{Hyper-parameter analysis of $\lambda_1$ and number of forwards on the PUMA panoptic segmentation. The metrics are averaged PQ across two semantic categories (SC) and three instance categories (IC).}
  \label{fig:visual_hyper}
  % \vspace{-1.0em}
\end{figure}

It is worth noting that the coefficient $\lambda_1$ in the loss function $\mathcal{L}_{\rm CoSeg}$ is crucial for Co-Seg++. For the possible value range as 0.1$\leq$$\lambda_1$$\leq$1, we perform a grid search at the panoptic segmentation setting on the PUMA dataset, as illustrated in the following Fig. \ref{fig:visual_hyper} (1). It is observed that our Co-Seg++ achieves the best performance with the average Dice of 53.10\% and 50.48\% when $\lambda_1$ value is set to 0.5. Moreover, we investigate the effect of the number of forwards on our Co-Seg++. From the illustration in Fig. \ref{fig:visual_hyper} (2), we observe that our Co-Seg++ achieves the best performance when the number of forwards is set to 2, where additional forward passes introduce excessive computational complexity, hindering the model's ability to effectively capture the complementary relationships between semantic and instance segmentation, resulting the performance degradation. In general, the appropriate coefficient $\lambda_1$ and the number of forwards can significantly improve the segmentation accuracy of our Co-Seg++ framework.

\begin{table*}[!t]
\centering
\small
\setlength\tabcolsep{4.4pt}
\caption{Evaluation on domain shift across different histopathology datasets.}
\scalebox{1}{\begin{tabular}{l|ccc|ccc|cc|cc|cc|cc}
\hline
& \multicolumn{6}{c|}{PUMA$\rightarrow$BCSS} & \multicolumn{2}{c|}{PUMA$\rightarrow$DSB} & \multicolumn{6}{c}{CRAG$\rightarrow$CoNSeP} \\
\cline{2-15}
Methods & \multicolumn{3}{c|}{Tumor} &  \multicolumn{3}{c|}{Stroma} &  \multicolumn{2}{c|}{Nucleus} &  \multicolumn{2}{c|}{Epithelial} &  \multicolumn{2}{c|}{LR} &  \multicolumn{2}{c}{Connective} \\
\cline{2-15}
& Dice$\uparrow$ & mIoU$\uparrow$ & HD$\downarrow$ & Dice$\uparrow$ & mIoU$\uparrow$ & HD$\downarrow$ & F1$\uparrow$ & AJI$\uparrow$ & F1$\uparrow$ & AJI$\uparrow$ & F1$\uparrow$ & AJI$\uparrow$ & F1$\uparrow$ & AJI$\uparrow$ \\
\hline
nnUnet \cite{isensee2021nnu}  & 26.43 & 19.15 & 582.37 & 23.58 & 16.42 & 545.28 & 41.26 & 39.13 & 42.37 & 34.16 & 29.54 & 19.82 & 18.43 & 15.67 \\
Swin-Umamba \cite{liu2024swin}  & 28.91 & 21.36 & 568.24 & 25.74 & 18.29 & 531.65 & 44.58 & 42.27 & 44.82 & 36.29 & 31.78 & 21.54 & 20.16 & 17.23 \\
EMCAD \cite{rahman2024emcad}  & 30.17 & 22.64 & 561.45 & 27.35 & 19.58 & 524.83 & 48.35 & 45.81 & 46.51 & 37.94 & 33.29 & 22.87 & 21.74 & 18.55 \\
H-SAM \cite{cheng2024unleashing}  & 32.20 & 23.63 & 550.16 & 29.10 & 20.88 & 510.35 & 63.45 & 59.39 & 52.79 & 41.87 & 38.64 & 27.45 & 27.91 & 24.25\\ 
Zig-RiR \cite{chen2025zig} & 31.48 & 23.17 & 556.72 & 28.26 & 20.14 & 518.94 & 58.87 & 56.96 & 49.65 & 39.83 & 37.92 & 26.47 & 24.38 & 21.16 \\
\hline
HoverNet \cite{graham2019hover} & 34.27 & 25.18 & 547.83 & 30.46 & 22.15 & 507.62 & 56.94 & 54.23 & 46.82 & 37.68 & 33.51 & 22.94 & 24.73 & 21.18 \\
HookNet \cite{van2021hooknet} & 32.65 & 24.01 & 555.29 & 29.38 & 21.27 & 516.47 & 51.82 & 48.74 & 43.15 & 34.82 & 35.68 & 24.92 & 22.57 & 19.38\\
CellPose \cite{stringer2021cellpose} & 36.28 & 27.15 & 542.68 & 32.17 & 23.86 & 502.91 & 45.81 & 46.49 & 44.29 & 36.17 & 35.82 & 24.58 & 25.16 & 21.47 \\
CPPNet \cite{chen2023cpp} & 35.19 & 26.24 & 546.35 & 30.59 & 22.25 & 501.40 & 53.75 & 43.41 & 50.27 & 40.18 & 38.74 & 27.16 & 25.31 & 21.83 \\
CellViT \cite{horst2024cellvit} & 36.85 & 27.59 & 541.27 & 32.74 & 24.38 & 501.15 & 58.72 & 55.89 & 48.13 & 39.52 & 34.62 & 23.71 & 23.25 & 20.91\\
PromptNucSeg \cite{shui2024unleashing} & 40.56 & 31.73 & 552.10 & 35.92 & 27.14 & 498.63 & 65.92 & 61.98 & 54.38 & 43.17 & 43.59 & 30.28 & 29.47 & 25.36 \\
PathoSAM \cite{griebel2025segment} & 43.72 & 34.86 & 525.43 & 37.15 & 28.23 & 489.36 & 69.27 & 64.15 & 55.81 & 43.92 & 44.52 & 31.09 & 32.18 & 27.63 \\
\hline
Co-Seg++ & \textbf{51.46} & \textbf{43.39} & \textbf{486.16} & \textbf{40.11} & \textbf{30.61} & \textbf{463.00} & \textbf{74.14} & \textbf{66.23} & \textbf{57.34} & \textbf{44.61} & \textbf{47.01} & \textbf{32.74} & \textbf{34.95} & \textbf{29.38}\\
\hline
\end{tabular}}
\label{tab:domain_shift}
\end{table*}

\begin{table}[!t]
\centering
\setlength\tabcolsep{2pt}
\caption{Impact of training data proportion on PUMA panoptic segmentation.}
\scalebox{0.95}{\begin{tabular}{c|l|cc|ccc}
\hline
\multirow{2}{*}{Training Data} & \multirow{2}{*}{Methods} & \multicolumn{2}{c|}{Semantic} & \multicolumn{3}{c}{Instance} \\ 
\cline{3-7}
& & Tumor & Stroma & Tumor & Lymph & Other Cells \\ 
\hline
\multirow{4}{*}{10\%} 
& CellViT \cite{horst2024cellvit} & 42.87 & 26.14 & 45.38 & 35.74 & 18.52 \\ 
& PromptNucSeg \cite{shui2024unleashing} & 43.68 & 26.92 & 45.96 & 36.47 & 19.26 \\ 
& PathoSAM \cite{griebel2025segment} & 41.56 & 25.38 & 44.17 & 34.29 & 17.84 \\ 
& Co-Seg++ & \textbf{46.05} & \textbf{29.78} & \textbf{48.61} & \textbf{43.48} & \textbf{20.73} \\ 
\hline
\multirow{4}{*}{30\%} 
& CellViT \cite{horst2024cellvit} & 50.16 & 31.05 & 53.29 & 42.81 & 22.15 \\ 
& PromptNucSeg \cite{shui2024unleashing} & 51.05 & 31.58 & 53.74 & 43.29 & 23.08 \\ 
& PathoSAM \cite{griebel2025segment} & 49.28 & 30.16 & 52.43 & 41.95 & 21.37 \\ 
& Co-Seg++ & \textbf{53.42} & \textbf{34.55} & \textbf{56.73} & \textbf{50.38} & \textbf{14.10} \\ 
\hline
\multirow{4}{*}{50\%} 
& CellViT \cite{horst2024cellvit} & 56.42 & 34.89 & 59.18 & 48.06 & 24.96 \\ 
& PromptNucSeg \cite{shui2024unleashing} & 57.12 & 35.24 & 59.38 & 48.67 & 25.41 \\ 
& PathoSAM \cite{griebel2025segment} & 56.89 & 35.21 & 60.15 & 48.52 & 24.85 \\ 
& Co-Seg++ & \textbf{59.18} & \textbf{38.26} & \textbf{61.48} & \textbf{53.91} & \textbf{21.73} \\ 
\hline
\multirow{4}{*}{100\%} 
& CellViT \cite{horst2024cellvit} & 61.95 & 38.28 & 64.72 & 52.81 & 27.69 \\ 
& PromptNucSeg \cite{shui2024unleashing} & 61.83 & 38.19 & 63.58 & 52.74 & 27.52 \\ 
& PathoSAM \cite{griebel2025segment} & 62.17 & 38.45 & 65.01 & 52.86 & 27.88 \\ 
& Co-Seg++ & \textbf{64.52} & \textbf{41.67} & \textbf{67.37} & \textbf{54.60} & \textbf{29.48} \\ 
\hline
\end{tabular}}
\label{tab:limited_anno}
\end{table}

\subsection{Computation Efficiency Analysis}
To evaluate the computation efficiency of Co-Seg++, we conducted comparisons with the independent segmentation model CellViT \cite{horst2024cellvit} and the encoder-shared segmentation model PathoSAM \cite{griebel2025segment} in terms of resource costs, including the FLOPs, inference latency, learnable and total parameters, as shown in Table \ref{tab:efficiency}. It is important to note that Co-Seg++ and PathoSAM can perform both semantic and instance segmentation through a single unified model, while CellViT \cite{horst2024cellvit} is task-specific and requires training two separate models to accomplish both segmentation tasks. Therefore, for CellViT \cite{horst2024cellvit}, we report the cumulative costs of two models. We observe that our Co-Seg++ demonstrates significant computation efficiency across all metrics. In particular, Co-Seg++ requires 869.67G FLOPs, representing a 5.6× reduction compared to the independent CellViT \cite{horst2024cellvit} and 3.2× reduction compared to the encoder-shared PathoSAM \cite{griebel2025segment}. In terms of inference latency, Co-Seg++ achieves 186.53ms per image, which is 4.4× lower than CellViT \cite{horst2024cellvit} and 2.4× lower than PathoSAM \cite{griebel2025segment}. Moreover, Co-Seg++ requires only 22.18M learnable parameters, representing a remarkable 40.2× reduction compared to CellViT \cite{horst2024cellvit} and 29.1× reduction compared to PathoSAM \cite{griebel2025segment}. Moreover, Co-Seg++ achieves superior performance, outperforming PathoSAM \cite{griebel2025segment} by 2.63\% (Dice) and 4.03\% (AJI) for semantic and instance segmentation, respectively, across three histopathology datasets. These comprehensive evaluations prove the superior computation efficiency of our Co-Seg++ framework in versatile medical segmentation.

\subsection{Domain Shift Assessment}
We further validate the robustness of Co-Seg++ under realistic domain variations. Specifically, we design three evaluation scenarios: (1) training on PUMA (Nanozoomer XR C12000 scanner) and testing on BCSS \cite{amgad2019structured} (\textit{i.e.}, sourced from TCGA \cite{weinstein2013cancer} database with multiple scanners) for tissue semantic segmentation, (2) training on PUMA (\textit{i.e.}, H\&E staining) and testing on DSB \cite{caicedo2019nucleus} (\textit{i.e.}, fluorescence staining) for nuclei instance segmentation, and (3) training on CRAG (\textit{i.e.}, Zeiss MIRAX MIDI scanner) and testing on CoNSeP \cite{graham2019hover} (\textit{i.e.}, Omnyx VL120 scanner) for multi-class nuclei segmentation. As shown in Table \ref{tab:domain_shift}, Co-Seg++ consistently outperforms all compared methods across different domain shift scenarios with statistical significance. For the PUMA$\rightarrow$BCSS cross-scanner evaluation, Co-Seg++ achieves 51.46\% Dice and 43.39\% mIoU for tumor segmentation, outperforming PathoSAM \cite{griebel2025segment} by 7.74\% and 8.53\%, while for stroma segmentation, Co-Seg++ attains 40.11\% Dice and 30.61\% mIoU with improvements of 2.96\% and 2.38\%. In the PUMA$\rightarrow$DSB cross-staining evaluation, Co-Seg++ demonstrates remarkable adaptability with 74.14\% F1 and 66.23\% AJI, surpassing PathoSAM \cite{griebel2025segment} by 4.87\% and 2.08\%, and H-SAM \cite{cheng2024unleashing} by 10.69\% and 6.84\%. For the CRAG$\rightarrow$CoNSeP task, Co-Seg++ achieves the highest performance across all cell types, with 57.34\% F1 for epithelial nuclei and 47.01\% F1 for lympho-reticular cells, representing gains of 1.53\% and 2.49\% over PathoSAM \cite{griebel2025segment}. Compared to classical methods like nnUNet \cite{isensee2021nnu}, Co-Seg++ demonstrates substantial improvements exceeding 20\% in Dice across multiple scenarios, validating that Co-Seg++ maintains superior generalization capabilities across diverse scanner types and staining protocols.

\subsection{Limited Annotation Analysis}
To investigate the stability of Co-Seg++ under limited supervision, we conduct sensitivity analysis by training our framework with varying proportions of training data (\textit{i.e.}, 10\%, 30\%, 50\%, and 100\%) on the PUMA dataset for panoptic segmentation. As presented in Table \ref{tab:limited_anno}, Co-Seg++ consistently outperforms state-of-the-art methods across all annotation scales for both semantic and instance segmentation tasks. With only 10\% of training data, Co-Seg++ achieves 46.05\% and 48.61\% panoptic quality for tumor semantic and instance segmentation, representing improvements of 2.37\% and 2.65\% over PromptNucSeg \cite{shui2024unleashing}, while for the challenging lymphocyte detection task, Co-Seg++ achieves 43.48\% compared to PromptNucSeg's 36.47\%, demonstrating a substantial 7.01\% improvement. When training data increases to 30\% and 50\%, Co-Seg++ maintains superiority with consistent improvements of 2-7\% across all categories, achieving 53.42\% and 59.18\% for tumor semantic segmentation at 30\% and 50\% data respectively, outperforming PromptNucSeg \cite{shui2024unleashing} by 2.37\% and 2.06\%. Even with full annotation availability, Co-Seg++ maintains performance advantages, achieving 64.52\% and 41.67\% for tumor and stroma semantic segmentation, representing improvements of 2.69\% and 3.22\% over PathoSAM \cite{griebel2025segment}. Compared to CellViT \cite{horst2024cellvit}, which employs multi-task learning without collaborative mechanisms, Co-Seg++ demonstrates more stable improvements across all annotation scales, with margins of 3.18\% and 7.74\% at 10\% data and 2.57\% and 1.79\% at 100\% data for tumor semantic and lymphocyte instance segmentation, indicating that the collaborative learning paradigm provides complementary benefits independent of annotation scale. These results demonstrate the robustness of our Co-Seg++ framework in realistic scenarios where comprehensive annotations are challenging to acquire.

\begin{figure}[!t]
  % \vspace{-4.3cm}
  \centering
  \includegraphics[width=1\linewidth]{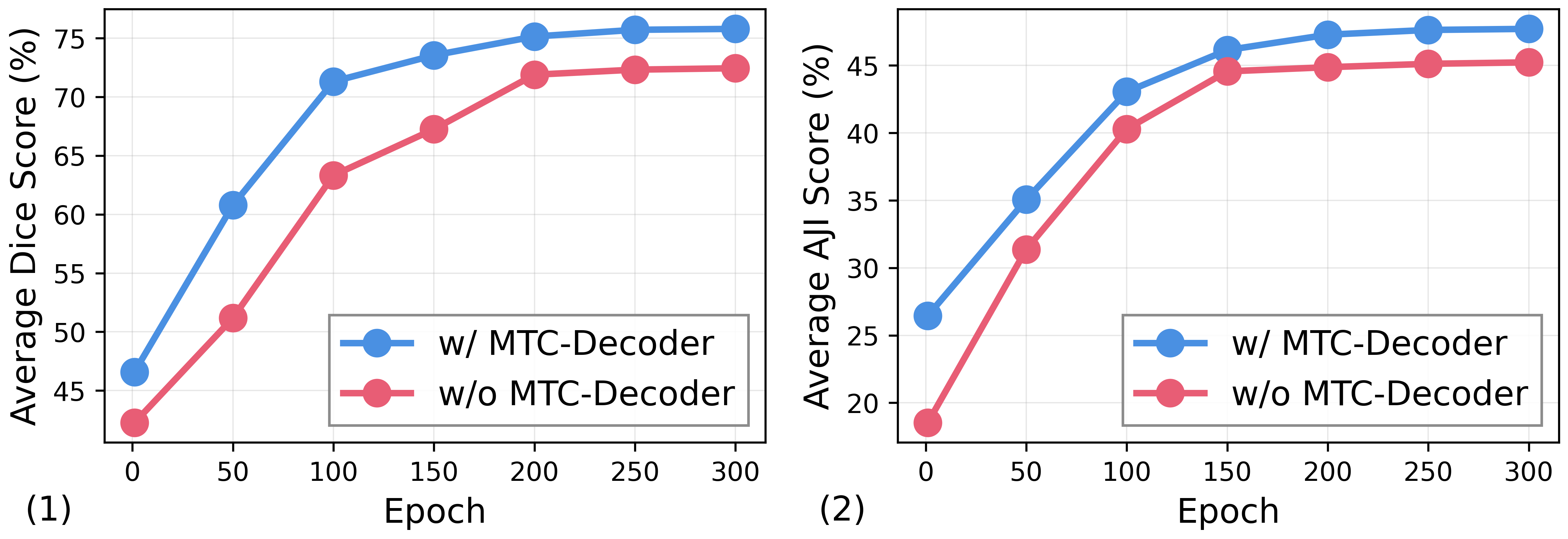}
  \caption{Training dynamics comparison of w/ MTC-Decoder and w/o MTC-Decoder on the PUMA dataset with two tasks, including (1) semantic and (2) instance segmentation.}
  \label{fig:decoder_analysis}
  % \vspace{-1.0em}
\end{figure}

\subsection{Interpretability Analysis of Mutual Guidance}
To analyze how cross-task guidance in the MTC-Decoder affects prediction dynamics and validate that mutual guidance consistently benefits both tasks, we conduct comprehensive training dynamics analysis and qualitative visualization. We compare two configurations on the PUMA dataset: with (w/) and without (w/o) MTC-Decoder, tracking task-specific performance after each training epoch to examine learning behavior over time. As illustrated in Fig. \ref{fig:decoder_analysis}, the configuration w/ MTC-Decoder consistently outperforms w/o MTC-Decoder across all training epochs for both semantic and instance segmentation tasks, with performance improvements of 2-3\% maintained from early training stages through convergence. Notably, both tasks benefit from the cross-task guidance mechanism without exhibiting negative transfer, where semantic segmentation achieves 64.52\% panoptic quality compared to 61.85\% w/o MTC-Decoder, and instance segmentation reaches 67.37\% compared to 64.58\% w/o MTC-Decoder.

\begin{figure}[!t]
  % \vspacedecoder_analysis
  \centering
  \includegraphics[width=0.98\linewidth]{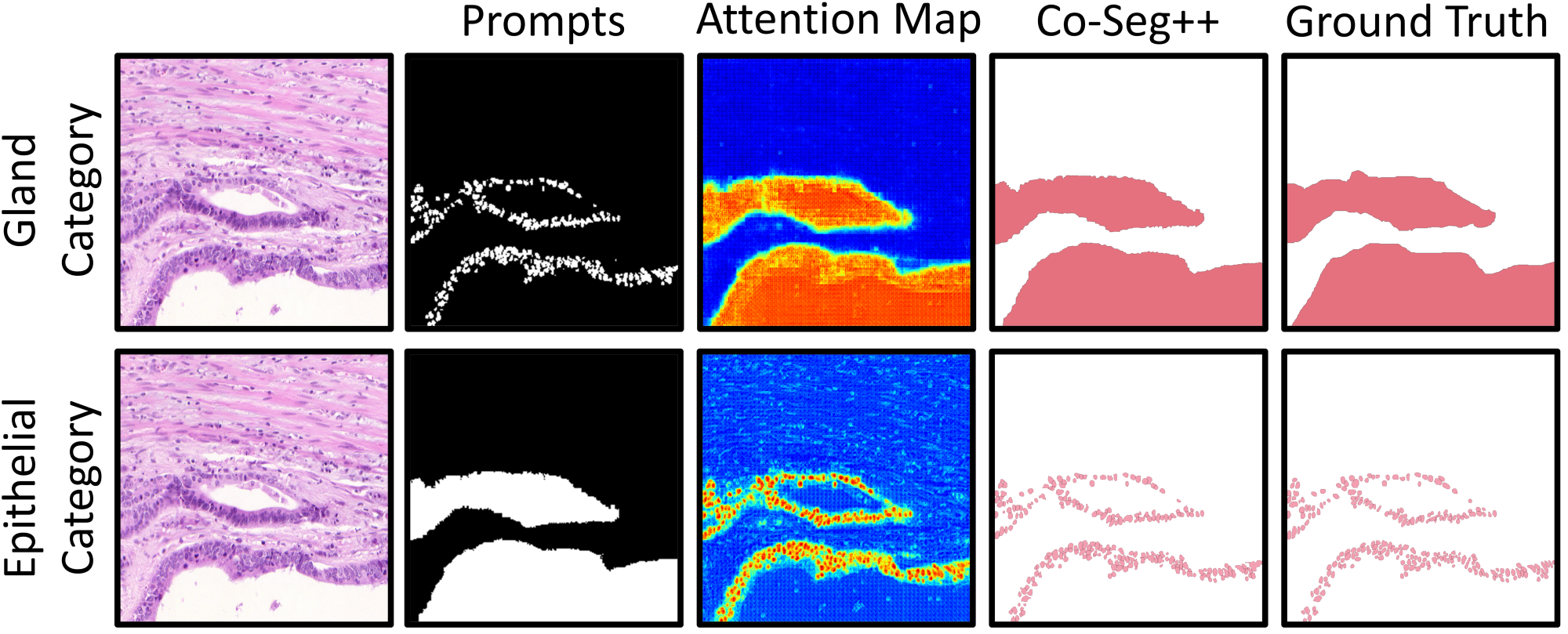}
  \caption{Visualization of cross-task guidance mechanism with binary mask prompts and attention maps on the GlaS dataset. Our Co-Seg++ with MTC-Decoder exhibits superior coherent semantic masks and accurate instance boundaries.}
  \label{fig:prompt_visual}
  % \vspace{-1.0em}
\end{figure}

Moreover, to further interpret how mutual guidance improves segmentation results, we perform qualitative visual analysis on the GlaS dataset for gland semantic segmentation and epithelial instance segmentation, as shown in Fig. \ref{fig:prompt_visual}. We visualize the binary mask prompts and cross-attention maps to demonstrate the information flow between tasks. Specifically, the semantic segmentation branch generates region-level prompts that capture global gland structure and contextual information, which effectively guides the instance branch to better distinguish individual cell boundaries within tissue regions by providing spatial constraints. Conversely, the instance segmentation branch produces fine-grained prompts highlighting precise object boundaries and morphological details, which in turn refine the semantic region boundaries and reduce over-segmentation errors. The attention maps reveal that when semantic features act as keys and values while instance features serve as queries, the cross-attention mechanism successfully integrates global contextual information into instance-level features, enabling the model to resolve ambiguous boundaries in densely packed or overlapping nuclei. Similarly, when instance features guide semantic segmentation, the fine-grained boundary information helps establish more precise tissue region delineation. These analyses validate that the MTC-Decoder enables both tasks to mutually enhance each other, where complementary contextual dependencies are effectively captured to improve segmentation consistency and accuracy.

\section{Conclusion}
In this work, we identify the interdependencies between semantic and instance segmentation in medical imaging, and propose a novel co-segmentation paradigm to establish the Co-Seg++ framework, allowing semantic and instance segmentation tasks to mutually enhance each other. The framework integrates a SSP-Encoder to capture long-range spatial and temporal relationships as prior constraints, and a MTC-Decoder that leverages cross-guidance to strengthen contextual consistency across tasks. Extensive experiments on diverse histopathology and CBCT datasets demonstrate that Co-Seg++ surpasses state-of-the-art methods across versatile segmentation, validating the effectiveness of our collaborative learning approach for comprehensive medical image understanding. Future work will focus on extending the co-segmentation paradigm to 3D volumetric medical image analysis, such as CT and MRI segmentation, and exploring its applicability to multi-modal medical imaging tasks.

\balance
\bibliographystyle{IEEEtran}
\bibliography{ref}

\end{document}